\documentclass[final]{article} 
\usepackage{iclr2027_conference,times}
\usepackage{url}
\usepackage{enumitem}
\usepackage{bbm}
\usepackage{lipsum}
\usepackage{graphicx}
\usepackage{tcolorbox}
\usepackage{soul}
\usepackage{caption}
\usepackage{subcaption}
\usepackage{wrapfig}
\usepackage{lipsum}
\usepackage{enumitem}
\usepackage{multirow}
\usepackage{tikz}
\usetikzlibrary{arrows.meta}
\usepackage{basic_commands}
\usepackage[endLComment=,italicComments=false]{algpseudocodex}
\usepackage{algorithm}
\usepackage{booktabs}
\usepackage{tabularew}
\usepackage{makecell}

\AddToHook{cmd/ttfamily/after}{\frenchspacing}

\definecolor{myred}{HTML}{F54254}
\definecolor{myorange}{HTML}{FFB135}
\definecolor{mygreen}{HTML}{10BD35}
\definecolor{myblue}{HTML}{BB2282}
\definecolor{plgray}{HTML}{999999}
\newcolumntype{R}[1]{>{\raggedleft\arraybackslash}p{#1}}
\newcolumntype{L}[1]{>{\raggedright\arraybackslash}p{#1}}
\renewcommand{\tt}[1]{\texttt{#1}}
\renewcommand{\phi}{\varphi}

\newcommand{\pl}[1]{{#1}}
\newcommand{\blue}{\textcolor{myblue}}
\newcommand{\methodname}{MAC-v2\xspace}
\newcommand{\achunk}{n}
\newcommand{\sg}{\mathrm{sg}}

\setlist[itemize]{itemsep=1pt, leftmargin=20pt}

\hypersetup{urlcolor=myblue,citecolor=myblue,linkcolor=myblue}

\title{MAC-v2: Long-horizon model-based value expansion from pixels}

\author{Kwanyoung Park  \quad Sergey Levine \\
    UC Berkeley \\
    \url{https://kwanyoungpark.github.io/MAC-v2/}
}

\iclrfinalcopy 
\begin{document}

\maketitle

\begin{abstract}
We study whether \emph{long-horizon model-based value expansion} can serve as an effective framework in pixel-based offline reinforcement learning (RL). Long-horizon model-based value expansion from pixels is substantially harder than state-based environments, as the agent must learn both the representation and its dynamics, and errors from either can accumulate recursively over imagined trajectories. 
We introduce \methodname, which enables long-horizon model-based expansion from pixels by training a reconstruction-free latent world model with action chunking. By learning representations from prediction and control objectives, \methodname does not need to encode control-irrelevant visual details that can hinder the dynamics prediction. Action chunking further mitigates the prediction error by reducing the number of recursive model predictions required for value expansion. Across diverse visual benchmarks including visual OGBench, MMBench2, Robomimic, and LIBERO-10, \methodname achieves strong performance against model-free and model-based offline RL baselines. Our analyses show that long-horizon model-based value expansion is critical for the performance of \methodname, and that both action chunking and latent learning objectives are essential for reliable long-horizon latent imagination. 
\end{abstract}

\section{Introduction}
Many real-world decision-making problems, such as robotic manipulation, require agents to learn directly from high-dimensional visual observations. This is particularly challenging in offline reinforcement learning (RL), where policies must be learned entirely from a fixed dataset without additional environment interaction. Unlike online RL, offline RL cannot evaluate the policy through environment interaction and rely on off-policy value estimates learned from the fixed dataset. As the learned policy deviates from the behavior policy, these off-policy estimates can become inaccurate due to distribution shift. Model-based value expansion~\citep{mve_feinberg2018} offers an appealing alternative by learning a world model, where on-policy trajectories can be generated by using a world model without accessing the real environment~\citep{mopo_yu2020,morel_kidambi2020,mobile_sun2023,dreamerv4_hafner2025,leq_park2025}.

Prior works have shown that model-based value expansion can be applied to offline RL from pixels, but typically only over short imagination horizons. For example, methods such as LOMPO~\citep{rafailov2021offline}, OfflineDV2~\citep{vd4rl_lu2023}, and SeMOPO~\citep{wan2024semopo} learn latent world models from visual observations and mitigate model exploitation through uncertainty estimation, conservative penalties, or representation decomposition. While these approaches demonstrate that model-based value expansion can be effective, their model rollouts are generally kept short (e.g., LOMPO, OfflineDV2, and SeMOPO use an imagination horizon of $5$ steps). In contrast, recent state-based offline RL methods have shown that substantially longer model-based value expansion can be effective for long-horizon tasks~\citep{mac_park2025,chung2026offline}.

Long-horizon model-based value expansion from pixels is substantially harder, as the agent must learn both the representation and its dynamics from the fixed offline data. Since pixel observations contain substantial control-irrelevant information~\citep{fu2026back,luo2026being}, training a world model directly on pixels becomes unnecessarily challenging.
This motivates learning a compact latent representation for world modeling. Here, the key challenge is to learn a representation that preserves the information needed for control, and also be predictable by the learned dynamics~\citep{islam2022agent,sun2024learning}. Otherwise, this can lead to inaccurate value estimation, as errors from either the representation or the dynamics can accumulate recursively over imagined trajectories, and this issue is further exacerbated in the offline setting, where the agent cannot collect corrective data when those trajectories reach poorly modeled regions.

We introduce \methodname, which enables long-horizon model-based value expansion from pixels by training a reconstruction-free latent world model with action chunking. By learning representations from prediction and control objectives, \methodname does not need to encode control-irrelevant visual details that can hinder the dynamics prediction, while also preserving the information needed for control. 
Moreover, we utilize action chunking~\citep{aloha_zhao2023} for the policy and the world model, reducing the number of recursive model predictions required over long horizons~\citep{mac_park2025}. 
Together, these design choices mitigate the compounding error during imagined rollouts and enable long-horizon model-based value expansion from visual observations. 

We evaluate \methodname across a broad suite of visual benchmarks and show that it matches or substantially outperforms prior model-free and model-based RL methods that rely on shorter-horizon value estimates. \methodname substantially outperforms prior model-free and model-based baselines on 25 visual OGBench~\citep{ogbench_park2025}, and matches the performance of the best prior methods in 10 MMBench2~\citep{Hansen2026Hallucination} tasks. We further evaluate \methodname on robotic manipulation benchmarks, including Robomimic~\citep{robomimic_mandlekar2021} and LIBERO-10~\citep{liu2023libero}, where the offline datasets consist primarily of expert demonstrations and behavior cloning policy serves as a strong baseline. On Robomimic, \methodname is the only RL method that matches or exceeds behavior cloning across all evaluated tasks, while remaining competitive with strong behavioral baselines on LIBERO-10. Our analyses show that long-horizon model-based value expansion is critical for this performance, and learning a reconstruction-free encoder with both world model and control objectives is important for reliable long-horizon latent rollouts. Overall, our results suggest that, with the design choices of \methodname, latent world models can support effective long-horizon model-based value expansion from pixels.

\section{Related Works}

\textbf{Offline model-free RL.}
Offline RL aims to learn policies solely from pre-collected datasets without further environment interaction~\citep{batch_lange2012,offline_levine2020}. A central challenge is distribution shift, as learned policies may select out-of-distribution states or actions for which value estimates are unreliable. Prior work addresses this issue through conservative objectives~\citep{cql_kumar2020}, behavior regularization~\citep{brac_wu2019,awr_peng2019,awac_nair2020,td3bc_fujimoto2021,rebrac_tarasov2023,fql_park2025}, in-sample value maximization~\citep{iql_kostrikov2022,sql_xu2023,xql_garg2023}, and sample-selection methods~\citep{sfbc_chen2023,idql_hansenestruch2023}. More recently, action chunking has shown strong performance on long-horizon offline RL tasks~\citep{qc_li2025,dqc_li2025,cgql_song2026} by improving temporal consistency and reducing the effective decision horizon. However, these model-free approaches rely on short-horizon off-policy targets for value learning, which can limit performance due to inaccurate value propagation~\citep{sharsa_park2025}. In contrast, \methodname learns values from long-horizon on-policy targets generated by a model through action-chunked latent rollouts, reducing the error from the recursive bootstrapping.

\textbf{Offline model-based RL.}
\methodname is a model-based offline RL method. Such methods learn a dynamics model from the fixed dataset and use it to generate synthetic trajectories for policy evaluation and optimization. By evaluating the policy through the learned model, they can utilize on-policy value estimation instead of off-policy estimates. However, policies may exploit inaccuracies in the learned dynamics model, leading to unreliable value estimates. Prior work addresses this issue through conservative objectives~\citep{morel_kidambi2020,combo_yu2021}, uncertainty-aware penalties~\citep{mopo_yu2020,rambo_rigter2022,mobile_sun2023,ravl_sims2024,leq_park2025}, and in-distribution policy optimization~\citep{mac_park2025}. In particular, MAC~\citep{mac_park2025} combines action-chunked dynamics with a rejection sampling policy, enabling long-horizon model-based value expansion on state-based tasks. \methodname builds on this framework, but addresses the additional challenge of learning a representation that can support accurate recursive model predictions directly from visual observations.

\textbf{Online RL from pixels.}
Learning from pixels requires representations that retain information relevant for control while remaining compact and predictable. Prior work has studied this problem extensively in online RL, including model-free approaches~\citep{dqn_mnih2013,drq_kostrikov2021,simba_lee2025} and model-based approaches~\citep{dreamer_hafner2020,dreamerv3_hafner2025,tdmpc_hansen2022,tdmpc2_hansen2024}. Representation objectives include image reconstruction~\citep{dreamer_hafner2020,dreamerv3_hafner2025,dreamerv4_hafner2025}, contrastive learning~\citep{laskin2020curl,stooke2021decoupling}, and reconstruction-free latent prediction~\citep{muzero_schrittwieser2020,tdmpc2_hansen2024,maes2026leworldmodel}. 
\methodname adopts the reconstruction-free latent prediction paradigm established by the prior works, and we do not propose a entirely new representation learning objective. 
Instead, we study how to utilize such representations for long-horizon model rollouts, required by offline model-based value expansion. We show that, when combined with action chunking, predictive latent representations can support long-horizon imagination and value estimation from a fixed visual dataset.

\textbf{Offline model-based RL from pixels.}
Several prior works have specifically studied model-based offline RL from visual observations using reconstruction-based latent representations. LOMPO~\citep{rafailov2021offline} learns a latent dynamics model and calculates model uncertainty in latent space, while OfflineDV2~\citep{vd4rl_lu2023} adapts DreamerV2~\citep{hafner2020dreamerv2} to offline visual control with uncertainty-aware penalties.
However, model imagination is typically kept relatively short  (e.g., LOMPO and OfflineDV2 use imagination of $5$ steps). We instead study whether predictive latent representations can support \emph{long-horizon model-based value expansion} from pixels. By combining latent prediction with action chunking, \methodname enables stable value expansion over more than $100$ environment steps.

\section{Preliminaries}

\textbf{Problem setting.}
We consider a block Markov decision process (MDP) defined as
$\gM=(\gS,\gO,\gA,r,\mu,p,q)$, where $\gS$ is the latent environment state space, $\gO$ is the observation space, and $\gA=\sR^d$ is the action space.
The reward function is $r(\pl{s},\pl{a}):\gS\times\gA\rightarrow\sR$,
$\mu(\pl{s})\in\Delta(\gS)$ is the initial state distribution,
$p(\pl{s}'\mid\pl{s},\pl{a}):\gS\times\gA\rightarrow\Delta(\gS)$ is the transition dynamics, and $q(\pl{o}\mid\pl{s}):\gS\rightarrow\Delta(\gO)$ is the observation model. The agent does not observe $s_t$ directly and instead receives $o_t\sim q(\cdot\mid s_t)$.
We assume that the current observation is sufficient to identify the underlying state for decision making, so that reactive policies $\pi(a_t\mid o_t)$ are sufficient. For a policy $\pi$, we define
$V^\pi(s)=\E_{\tau\sim p^\pi(\tau\mid s_0=s)}
\left[\sum_{t=0}^{\infty}\gamma^t r(s_t,a_t)\right],$
where $\gamma\in(0,1)$ is the discount factor. The goal of offline RL is to learn a policy maximizing $\E_{s_0\sim\mu}[V^\pi(s_0)]$ using only a fixed dataset $\gD=\{\tau^{(i)}\}$ of previously collected trajectories, without additional environment interaction.

\paragraph{Flow Matching Policy.}
A flow matching policy~\citep{flow_lipman2023,flow_albergo2023,flow_liu2023} represents a noise-conditioned deterministic policy
$a=\pi_\theta(z,\epsilon)$
by transporting samples from a simple reference distribution to the action distribution conditioned on a state.
Let
$v_\theta:[0,1]\times\gZ\times\gA\rightarrow\gA$
denote a state-conditioned, time-dependent velocity field, where $u\in[0,1]$ denotes the flow time.
Given an initial noise sample $\epsilon\sim\gN(0,I_d)$, the corresponding flow
$\psi_\theta(u;z,\epsilon)$
is defined by the ordinary differential equation (ODE)~\citep{smooth_lee2002}
\begin{align}
    \frac{\de}{\de u}\psi_\theta(u;z,\epsilon)
    =
    v_\theta\!\left(u,z,\psi_\theta(u;z,\epsilon)\right),
    \qquad
    \psi_\theta(0;z,\epsilon)=\epsilon.
\end{align}
The policy is defined as the endpoint of this flow,
\begin{align}
    \pi_\theta(z,\epsilon)
    =
    \psi_\theta(1;z,\epsilon),
\end{align}
which induces a conditional action distribution $\pi_\theta(a\mid z)$ through
$\epsilon\sim\gN(0,I_d)$.

Given state-action samples $(s,a)\sim\gD$, flow matching trains the velocity field along linear interpolation paths between Gaussian noise $\epsilon$ and the target action $a$.
Specifically, for
\begin{align}
    a_u=(1-u)\epsilon+ua,
\end{align}
the velocity field is trained by minimizing
\begin{align}
    L^\mathrm{flow}(\theta)
    =
    \E_{\substack{
        (s,a)\sim\gD,\,
        \epsilon\sim\gN(0,I_d),\\
        u\sim\mathrm{Unif}([0,1])
    }}
    \left[
        \left\|
        v_\theta(u,s,a_u)
        -
        (a-\epsilon)
        \right\|_2^2
    \right].
    \label{eq:flow}
\end{align}
After training, actions are sampled by drawing
$\epsilon\sim\gN(0,I_d)$
and numerically integrating the learned velocity field from $u=0$ to $u=1$.

\section{\methodname}

We present \methodname, an offline model-based RL framework that enables \emph{long-horizon model-based value expansion} from visual observations. 
\methodname builds on two lines of work: model-based value expansion with action chunking from MAC~\citep{mac_park2025}, and latent world models trained with prediction and control objectives akin to reconstruction-free online model-based RL literatures~\citep{muzero_schrittwieser2020,tdmpc_hansen2022,tdmpc2_hansen2024}.
Neither line of work alone enables long-horizon model-based value expansion in pixel-based offline RL, as \methodname assumes low-dimensional state observations, while prior reconstruction-free latent world models have been limited to short-horizon rollouts.

Our key design choice is to learn a reconstruction-free latent dynamics with action chunking. With action chunking size of $\achunk$, an imagination horizon of $H\achunk$ environment steps requires only $H$ recursive model predictions, reducing the compounding of prediction errors. Moreover, we learn the representation with world modeling and behavior cloning objective, encouraging the representation to preserve the information for reward prediction and control, while being predictable under the learned dynamics model. We describe our world model objectives in \Cref{sec:repr}, and the long-horizon model-based value expansion with the learned latent space in \Cref{sec:mve}.

\subsection{Learning Representations for Long-Horizon Imagination}
\label{sec:repr}
We encode each image observation into a latent state using a deterministic encoder
$f_\xi:\gO\rightarrow\gZ$.
Rather than reconstructing the input image, \methodname learns the encoder through objectives relevant to latent prediction and control. In particular, gradients from the dynamics, reward, and behavior cloning objectives are propagated into the encoder, learning a compact representation that retains information for predicting future latent states, rewards, and control.

Concretely, given a latent state and an action chunk, we learn a deterministic dynamics model
$p_\psi:\gZ\times\gA^\achunk\rightarrow\gZ$
to predict the latent state after $\achunk$ environment steps, by minimizing
\begin{align}
    L^\mathrm{dyn}(\psi, \xi) &= \E_{(o_t, a_t, \cdots, o_{t+\achunk}) \sim \gD} \left[\left\|p_\psi(f_{\xi}(o_t), \bm{a}_t) - \sg(f_{\xi}(o_{t+\achunk})) \right\|_2^2 \right],
    \label{eq:dyn}
\end{align}
where $\sg(\cdot)$ denotes the stop-gradient operator.

We also learn a distributional action-chunked reward model
$r_\psi:\gZ\times\gA^\achunk\rightarrow\Delta(\mathbb{R})$ with two-hot encoding~\citep{dreamerv3_hafner2025}. We regress the target return $\bm r_t = \sum_{i=0}^{n-1} \gamma^{i} r_{t+i}$ by minimizing 
\begin{align}
L^\mathrm{rew}(\psi, \xi) &= \E_{(o_t, \bm{a}_t) \sim \gD} \left[ -\log r_\psi(\bm{r}_t | f_{\xi}(o_t), \bm{a}_t) \right]. \label{eq:rew}
\end{align}

We parameterize the behavioral policy as a conditional flow model, and train a state-conditioned velocity field
$v_\theta:\sR \times \gZ \times \gA^\achunk \rightarrow \gA^\achunk$
using the flow-matching objective
\begin{align}
    L^\mathrm{flow}(\theta, \xi) &= \E_{\substack{\epsilon \sim \gN(0, I_{\achunk d}), \ (o_t, \bm{a}_{t}) \sim \gD, \\ u \sim \mathrm{Unif}([0, 1]), \ \bm{a}_\epsilon = (1-u)\epsilon+u\bm{a}_{t}}}\left[\| v_\theta(u, f_\xi(o_t), \bm{a}_\epsilon) - (\bm{a}_t - \epsilon) \|_2^2\right].
    \label{eq:flow_ac}
\end{align}
We denote by $\pi_\theta(z_t,\epsilon)$ the endpoint at $u=1$ obtained by integrating the learned velocity field from the initial noise $\epsilon \sim \gN(0, I_{nd})$.

\subsection{Long-Horizon Latent Value Expansion with Action Chunking}
\label{sec:mve}

With the learned latent world model, we apply the action-chunked value-learning framework of MAC~\citep{mac_park2025} in the latent space.
First, we extract the policy from the behavioral policy by  sampling behavioral action candidates $\hat{a}^{(i)}=\pi_\theta(z, \epsilon^{(i)})$ from multiple noises $\epsilon^{(i)} \sim \gN(0, I_{nd})$, and extracting the best action among the candidates via rejection sampling~\citep{sfbc_chen2023}:
\begin{align}
    \pi_\theta^*(z) &= \text{argmax}_{\hat a^{(1)}, \cdots, \hat a^{(M)}} Q_{\phi} (z, \hat a^{(i)}).
    \label{eq:rs}
\end{align}

\begin{figure}[t]
\vspace{-1.5em}
\begin{algorithm}[H]
\caption{\methodname}
\label{alg:training}
\begin{algorithmic}
    \Require Dataset $\mathcal{D}$, rollout length $H$, action chunking size $n$, rejection sampling size $M$,  \\
    encoder $f_{\xi}$, dynamics / reward model $p_{\psi}$ / $r_{\psi}$, 
    value $V_{\phi}$, Q-function $Q_{\phi}$,
    flow BC policy $\pi_{\theta}$ 

    \State
    \State \blue{\emph{// Training loop}}
    \While{not converged}
        \State \Comment{Sample action-chunked batch from the dataset ($\bm{a}_t = a_{t:t+n-1}$, $\bm{r}_t = \sum_{i=0}^{\achunk-1} \gamma^i r_{t+i}$)}
        \State Sample batch $\{(o_t, \bm{a}_{t}, \bm{r}_{t},  o_{t+\achunk})\} \sim \mathcal{D}$ 

        \State
        \State \Comment{Train \textbf{dynamics}, \textbf{reward} model (and \textbf{encoder}) using dataset transitions}
        \State Update $p_\psi, f_{\xi}$ to minimize 
        $\E [\|p_\psi(f_{\xi}(o_t), \bm{a}_t) - \sg(f_{\xi}(o_{t+\achunk}))\|_2^2]$ (\Cref{eq:dyn}) 
        \State Update $r_\psi, f_{\xi}$ to maximize 
        $\E [\log(\bm{r}_t \mid r_\psi(f_{\xi}(o_t), \bm{a}_t))]$ (\Cref{eq:rew}) 

        \State
        \State \Comment{Train \textbf{BC policy} (and \textbf{encoder}) using dataset transitions}
        \State Update $\pi_\theta, f_{\xi}$ with flow-matching loss (\Cref{eq:flow_ac})

        \vspace{0.2em}
        \State
        \State \Comment{Generate \textbf{model rollouts} in latent space ($\hat{z}_t = f_{\xi}(o_t)$)}
        \For{$k=0, 1, \ldots, H-1$}
            \State $\bm{\hat{a}}_{t+kn} \leftarrow ~$\textsc{policy}$(\hat{z}_{t+kn})$
            \State $\hat{z}_{t+(k+1)\achunk}, \bm{\hat{r}}_{t+kn} \sim p_{\psi}(\cdot \mid \hat{z}_{t+kn}, \bm{\hat{a}}_{t+kn})$, $r_{\psi}(\cdot \mid \hat{z}_{t+kn}, \bm{\hat{a}}_{t+kn})$
        \EndFor

        \State
        \State \Comment{Update \textbf{value} and \textbf{critic} using model rollouts}
        \State Update $V_\phi$ with $nH$-step targets from the rollout (\Cref{eq:loss_v})
        
        \State Update $Q_\phi$ with the learned value function $V_{\phi}$ (\Cref{eq:loss_q})
    \EndWhile

    \State
    \State \blue{\emph{// Extract action from flow BC policy $\pi_{\theta}$ with rejection sampling}}
    \Function{policy}{$z$}
        \State $\epsilon^{(i)} \sim \mathcal{N}(0, I)$
        \State $\bm{\hat{a}}^{(i)} = \pi_{\theta}(z, \epsilon^{(i)})$
        
        \Return $\text{argmax}_{\bm{\hat{a}}^{(1)}, \cdots, \bm{\hat{a}}^{(M)}} Q_{\phi} (z, \bm{\hat{a}}^{(i)})$
    \EndFunction
\end{algorithmic}
\end{algorithm}
\vspace{-2em}
\end{figure}

We train the value function purely from latent on-policy model rollout, generated as 
\begin{align}
    \mathcal{T} = (z_t, \bm{\hat{a}}_{t}, \bm{\hat r}_{t}, \hat z_{t+\achunk}, \bm{\hat{a}}_{t+\achunk}, \bm{\hat r}_{t+\achunk}, \ldots, \hat z_{t+H\achunk}),
\end{align}
where $\bm{\hat{a}}_{t}$ denotes the action chunk $\hat{a}_{t:t+n-1}$ generated from the rejection sampling policy $\pi_\theta^*(z_t)$, and $\bm{\hat{r}}_{t}$ denotes the discounted sum of rewards $\sum_{i=0}^{n-1} \gamma^i r_{t+i}$ predicted from the reward model $r_{\psi}(\cdot | z_t, \bm{a}_t)$.
Here, initial states $z_t$ is uniformly sampled from the dataset and encoded with $f_{\xi}$,
and subsequent actions, rewards, and next latent states are synthesized by our rejection-sampling policy, reward model, and dynamics model, respectively, where we denote with a hat notation.

After collecting the trajectory, we learn a distributional value function
$V_\phi(\pl{z_t}): \gZ \to \Delta(\sR)$
with the following loss:
\begin{align}
L^V(\phi) &= \E\left[-\log 
V_\phi(\hat v_{t+kn}|\hat{z}_{t+kn})
\right]; ~\hat v_{t+kn} = \sum_{i=k}^{H-1} \gamma^{(i-k)n} \bm{\hat r}_{t+in} + \gamma^{(H-k)n}v_{\bar \phi}(\hat z_{t+Hn})\label{eq:loss_v} 
\end{align}
where $\bar \phi$ denotes exponentially averaged target parameters~\citep{dqn_mnih2013}, $v_{\phi}(\cdot) = \E_{v \sim V_{\phi}(\cdot)}[v]$,
and the expectations are over 
$\mathcal{T} = (z_t=\hat z_t, \bm{\hat a}_t, \bm{\hat r}_t, \ldots, \hat z_{t+Hn})$ and $k$ uniformly sampled from $\{0, 1, \ldots, H-1\}$.

Finally, we train the action-chunked Q function 
$Q_\phi(\pl{z_t}, \pl{\bm{a}_{t}}): \gZ \times \gA^n \to \Delta(\sR)$ for the rejection sampling
with the following loss:
\begin{align}
L^Q(\phi) &= \E_{o_t \sim \gD} \left[-\log 
Q_\phi(\bm{\hat r}_{t} + \gamma^n v_\phi(\hat z_{t+n}) | z_{t}, \bm{\hat a}_{t})\right]. \label{eq:loss_q}
\end{align}
For both value learning, we use two-hot encoding for the distribution.
We provide a pseudocode for \methodname in \Cref{alg:training}.

\section{Experiments}

We now empirically evaluate \methodname and examine whether long-horizon model-based value expansion is effective for offline RL from pixels. First, we show that \methodname can serve as an effective general-purpose offline RL algorithm from visual observations in \Cref{sec:exp_vis}, by evaluating on offline RL benchmarks with diverse control problems and challenging long-horizon tasks where accurate value propagation is important. Second, we examine whether \methodname can be also applied in robotic manipulation settings, where the offline datasets are closer to expert demonstrations and strong behavioral cloning policies already provide competitive performance, in \Cref{sec:exp_rob}. Finally, we analyze the importance of long-horizon model-based value expansion of \methodname and ablate the training objectives used for world model learning in \Cref{sec:exp_qna}. Unless otherwise specified, all results are averaged over four random seeds.

\begin{table}[t!]
\caption{
\footnotesize
\textbf{Results on standard offline RL visual benchmark tasks.}
We report the success rate for OGBench tasks and normalized scores ($0$-$100$) for MMBench2 tasks across 4 seeds, averaged over 15 episodes. Best scores for each task are highlighted in \color{myblue}{magenta}.
}
\label{table:standard}

\vspace{-0.5em}
\centering
\scalebox{0.745}
{

\begin{tabular*}{1.33\textwidth}{p{5.6cm} R{0.5cm}@{\hspace{0.05cm}}L{0.33cm} R{0.5cm}@{\hspace{0.05cm}}L{0.33cm} R{0.5cm}@{\hspace{0.05cm}}L{0.33cm} R{0.5cm}@{\hspace{0.05cm}}L{0.33cm} R{0.5cm}@{\hspace{0.05cm}}L{0.33cm} R{0.5cm}@{\hspace{0.05cm}}L{0.33cm} R{0.5cm}@{\hspace{0.05cm}}L{0.33cm}
R{0.5cm}@{\hspace{0.05cm}}L{0.33cm}}
\toprule
\multicolumn{1}{c}{} & \multicolumn{10}{c}{\tt{Model-Free}} & \multicolumn{6}{c}{\tt{Model-Based}} \\
\cmidrule(r){2-11}
\cmidrule(r){12-17}
\multicolumn{1}{l}{\tt{Environment}} & \multicolumn{2}{c}{\makebox[1.15cm]{\tt{IQL}}} & \multicolumn{2}{c}{\makebox[1.15cm]{\tt{ReBRAC}}} & \multicolumn{2}{c}{\makebox[1.15cm]{\tt{IFQL}}} & \multicolumn{2}{c}{\makebox[1.15cm]{\tt{FQL}}} & \multicolumn{2}{c}{\makebox[1.15cm]{\tt{QC}}} & \multicolumn{2}{c}{\makebox[1.15cm]{\tt{LOMPO}}} & \multicolumn{2}{c}{\makebox[1.15cm]{\tt{OffDV2}}} & \multicolumn{2}{c}{\makebox[1.15cm]{\color{myblue}\tt{\methodname}}} \\
\midrule
\multicolumn{15}{l}{\texttt{OGBench (25 tasks)}}\\
\midrule
\tt{visual-cube-single (5 tasks)} & \tt{65} & {\tiny $\pm$\tt{4}} & \tt{74} & {\tiny $\pm$\tt{1}} & \tt{36} & {\tiny $\pm$\tt{5}} & \tt{62} & {\tiny $\pm$\tt{4}} & \tt{\color{myblue}{78}} & {\tiny $\pm$\tt{2}} & \tt{0} & {\tiny $\pm$\tt{0}} & \tt{0} & {\tiny $\pm$\tt{0}} & \tt{74} & {\tiny $\pm$\tt{1}} \\
\tt{visual-cube-double (5 tasks)} & \tt{13} & {\tiny $\pm$\tt{1}} & \tt{4} & {\tiny $\pm$\tt{3}} & \tt{2} & {\tiny $\pm$\tt{1}} & \tt{5} & {\tiny $\pm$\tt{3}} & \tt{3} & {\tiny $\pm$\tt{1}} & \tt{0} & {\tiny $\pm$\tt{0}} & \tt{0} & {\tiny $\pm$\tt{0}} & \tt{\color{myblue}{53}} & {\tiny $\pm$\tt{3}} \\
\tt{visual-scene (5 tasks)} & \tt{18} & {\tiny $\pm$\tt{8}} & \tt{33} & {\tiny $\pm$\tt{4}} & \tt{21} & {\tiny $\pm$\tt{2}} & \tt{41} & {\tiny $\pm$\tt{2}} & \tt{57} & {\tiny $\pm$\tt{2}} & \tt{2} & {\tiny $\pm$\tt{1}} & \tt{4} & {\tiny $\pm$\tt{2}} & \tt{\color{myblue}{72}} & {\tiny $\pm$\tt{12}} \\
\tt{visual-puzzle-3x3 (5 tasks)} & \tt{2} & {\tiny $\pm$\tt{3}} & \tt{19} & {\tiny $\pm$\tt{1}} & \tt{21} & {\tiny $\pm$\tt{1}} & \tt{22} & {\tiny $\pm$\tt{1}} & \tt{20} & {\tiny $\pm$\tt{0}} & \tt{3} & {\tiny $\pm$\tt{2}} & \tt{2} & {\tiny $\pm$\tt{1}} & \tt{\color{myblue}{95}} & {\tiny $\pm$\tt{4}} \\
\tt{visual-puzzle-4x4 (5 tasks)} & \tt{0} & {\tiny $\pm$\tt{0}} & \tt{13} & {\tiny $\pm$\tt{1}} & \tt{4} & {\tiny $\pm$\tt{5}} & \tt{10} & {\tiny $\pm$\tt{1}} & \tt{20} & {\tiny $\pm$\tt{2}} & \tt{0} & {\tiny $\pm$\tt{0}} & \tt{1} & {\tiny $\pm$\tt{0}} & \tt{\color{myblue}{46}} & {\tiny $\pm$\tt{5}} \\
\midrule
\multicolumn{15}{l}{\texttt{MMBench2 (10 tasks)}}\\
\midrule
\tt{dmcontrol (2 tasks)} & \tt{19} & {\tiny $\pm$\tt{2}} & \tt{39} & {\tiny $\pm$\tt{2}} & \tt{17} & {\tiny $\pm$\tt{2}} & \tt{\color{myblue}{53}} & {\tiny $\pm$\tt{1}} & \tt{22} & {\tiny $\pm$\tt{2}} & \tt{14} & {\tiny $\pm$\tt{4}} & \tt{44} & {\tiny $\pm$\tt{12}} & \tt{31} & {\tiny $\pm$\tt{4}} \\
\tt{maniskill (1 task)} & \tt{5} & {\tiny $\pm$\tt{2}} & \tt{8} & {\tiny $\pm$\tt{3}} & \tt{11} & {\tiny $\pm$\tt{4}} & \tt{3} & {\tiny $\pm$\tt{2}} & \tt{3} & {\tiny $\pm$\tt{2}} & \tt{\color{myblue}{23}} & {\tiny $\pm$\tt{3}} & \tt{0} & {\tiny $\pm$\tt{0}} & \tt{4} & {\tiny $\pm$\tt{2}} \\
\tt{ogbench (2 tasks)} & \tt{7} & {\tiny $\pm$\tt{4}} & \tt{15} & {\tiny $\pm$\tt{2}} & \tt{10} & {\tiny $\pm$\tt{3}} & \tt{30} & {\tiny $\pm$\tt{4}} & \tt{6} & {\tiny $\pm$\tt{2}} & \tt{0} & {\tiny $\pm$\tt{0}} & \tt{1} & {\tiny $\pm$\tt{0}} & \tt{\color{myblue}{33}} & {\tiny $\pm$\tt{4}} \\
\tt{pygame (5 tasks)} & \tt{41} & {\tiny $\pm$\tt{2}} & \tt{50} & {\tiny $\pm$\tt{5}} & \tt{47} & {\tiny $\pm$\tt{1}} & \tt{58} & {\tiny $\pm$\tt{1}} & \tt{50} & {\tiny $\pm$\tt{1}} & \tt{31} & {\tiny $\pm$\tt{3}} & \tt{43} & {\tiny $\pm$\tt{1}} & \tt{\color{myblue}{66}} & {\tiny $\pm$\tt{2}} \\

\bottomrule
\end{tabular*}
}
\vspace{-1.0em}
\end{table}

\subsection{Visual offline RL benchmarks}

\label{sec:exp_vis}
We first evaluate whether \methodname can serve as a general-purpose offline RL method from pixels. We consider two complementary benchmark suites. Visual OGBench~\citep{ogbench_park2025} contains structured manipulation problems with long temporal horizons, suited for evaluating whether model-based value expansion can propagate useful information over long sequences. MMBench2~\citep{Hansen2026Hallucination} provides a more diverse visual control tasks and diverse source of data collection, testing whether it can be used for beyond the structured manipulation setting.

\subsubsection{Experimental setting}

\textbf{Tasks and datasets.}
We evaluate on $25$ visual single-task environments from OGBench~\citep{ogbench_park2025} and report average success rates within each task family, following the evaluation protocol of \citet{fql_park2025}.
We additionally evaluate on $10$ ``unseen'' tasks from MMBench2~\citep{Hansen2026Hallucination}, which cover a diverse set of visual control problems. We resize observations to $64\times64$ and combine all the datasets (e.g., collected from curiosity-driven policy, expert policy, random policy) assigned to each task. For MMBench2, we report normalized return on a $0$-$100$ scale using the minimum and maximum trajectory returns from the benchmark.

\textbf{Methods.}
We compare \methodname against strong model-free and model-based offline RL baselines. For model-free methods, we consider IQL~\citep{iql_kostrikov2022}, ReBRAC~\citep{rebrac_tarasov2023}, IFQL, the flow-based variant of IDQL~\citep{idql_hansenestruch2023}, and FQL~\citep{fql_park2025}. 
For model-based methods, we focus specifically on prior \emph{offline} RL methods that learn from visual observations. This literature is considerably smaller than the broader literature on online visual model-based RL methods, which are not directly comparable in our setting. For offline model-based methods, we include LOMPO~\citep{rafailov2021offline}, which learns a latent world model and apply uncertainty-aware policy optimization on top of the frozen representation, OfflineDV2~\citep{vd4rl_lu2023} (denoted as OffDV2), which combines uncertainty penalty~\citep{mopo_yu2020} on top of the world model from DreamerV2~\citep{hafner2020dreamerv2}.  

\subsubsection{Results}
Table~\ref{table:standard} summarizes the results on the visual OGBench and MMBench2 tasks. On OGBench, \methodname outperforms all baselines in 4 out of 5 task families, and matches QC on the remaining one (\tt{visual-cube-single}). 
Notably, \methodname shows 95\% score in \tt{puzzle-3x3}, where the best prior method scores 22\%, while prior model-based RL methods score near-zero performance across all tasks.
On MMBench2, \methodname achieves the best performance on the \tt{ogbench} and \tt{pygame} domains, and remains competitive with strong model-free baselines such as FQL on the remaining tasks. Together, these results shows the effectiveness of long-horizon model-based value expansion across diverse visual offline RL problems.

\begin{table}[t!]
\caption{
\footnotesize
\textbf{Results on visual robotic benchmarks.}
We report the success rate across 4 seeds, averaged over 15 episodes. Best scores for each task are highlighted in \color{myblue}{magenta}.
}
\label{table:robotics}

\vspace{-0.5em}
\centering
\scalebox{0.76}
{

\begin{tabular*}{1.31\textwidth}{p{8.5cm} R{0.5cm}@{\hspace{0.05cm}}L{0.33cm} R{0.5cm}@{\hspace{0.05cm}}L{0.33cm} R{0.5cm}@{\hspace{0.05cm}}L{0.33cm} 
R{0.5cm}@{\hspace{0.05cm}}L{0.33cm} 
R{0.5cm}@{\hspace{0.05cm}}L{0.33cm} R{0.5cm}@{\hspace{0.05cm}}L{0.33cm} }
\toprule
\multicolumn{1}{c}{} & \multicolumn{6}{c}{\tt{Model-Free}} & \multicolumn{6}{c}{\tt{Model-Based}} \\
\cmidrule(r){2-7}
\cmidrule(r){8-13}
\multicolumn{1}{l}{\tt{Environment}} & \multicolumn{2}{c}{\makebox[1.15cm]{\tt{BC}}} & \multicolumn{2}{c}{\makebox[1.15cm]{\tt{QC}}} & \multicolumn{2}{c}{\makebox[1.15cm]{\tt{QC-IFQL}}} & \multicolumn{2}{c}
{\makebox[1.15cm]{\tt{LOMPO}}} & \multicolumn{2}{c}
{\makebox[1.15cm]{\tt{OffDV2}}} & \multicolumn{2}{c}{\makebox[1.15cm]{\color{myblue}\tt{\methodname}}} \\
\midrule
\multicolumn{9}{l}{\texttt{Robomimic (3 tasks)}}\\
\midrule
\tt{lift} & \tt{\color{myblue}{100}} & {\tiny $\pm$\tt{0}} & \tt{\color{myblue}{100}} & {\tiny $\pm$\tt{0}} & \tt{\color{myblue}{100}} & {\tiny $\pm$\tt{0}} & \tt{27} & {\tiny $\pm$\tt{42}} & \tt{51} & {\tiny $\pm$\tt{19}} & \tt{\color{myblue}{100}} & {\tiny $\pm$\tt{0}} \\
\tt{can} & \tt{\color{myblue}{97}} & {\tiny $\pm$\tt{3}} & \tt{\color{myblue}{97}} & {\tiny $\pm$\tt{3}} & \tt{92} & {\tiny $\pm$\tt{3}} & \tt{0} & {\tiny $\pm$\tt{0}} & \tt{0} & {\tiny $\pm$\tt{0}} & \tt{95} & {\tiny $\pm$\tt{3}} \\
\tt{square} & \tt{68} & {\tiny $\pm$\tt{21}} & \tt{55} & {\tiny $\pm$\tt{13}} & \tt{55} & {\tiny $\pm$\tt{10}} & \tt{0} & {\tiny $\pm$\tt{0}} & \tt{0} & {\tiny $\pm$\tt{0}} & \tt{\color{myblue}{72}} & {\tiny $\pm$\tt{10}} \\
\midrule
\multicolumn{9}{l}{\texttt{LIBERO (10 tasks)}}\\
\midrule
\tt{LIBERO-10 (10 tasks)} & \tt{\color{myblue}{80}} & {\tiny $\pm$\tt{1}} & \tt{77} & {\tiny $\pm$\tt{1}} & \tt{78} & {\tiny $\pm$\tt{3}} & \tt{1} & {\tiny $\pm$\tt{2}} & \tt{0} & {\tiny $\pm$\tt{0}} & \tt{\color{myblue}{80}} & {\tiny $\pm$\tt{1}} \\
\bottomrule
\end{tabular*}
}
\vspace{-1.0em}
\end{table}

\subsection{Robotic manipulation benchmarks}
\label{sec:exp_rob}

We next study the case where offline datasets are composed primarily of successful or near-expert demonstrations. In these datasets, behavioral cloning is already a strong baseline, which is challenging for offline RL algorithms to improve upon. This evaluation tests whether model-based value expansion can also be effective in these datasets, and whether it outperforms behavioral cloning policies.

\subsubsection{Experimental setting}

\textbf{Tasks and datasets.}
We evaluate on three visual manipulation tasks from Robomimic~\citep{robomimic_mandlekar2021} and the ten long-horizon tasks from LIBERO-10~\citep{liu2023libero}. For Robomimic, we use the multi-human (\tt{mh}) datasets and $84\times84$ observations from both the agent-view and wrist cameras. For LIBERO-10, we use the original $128\times128$ observations and only use the agent-view camera. We report average task success rates on both benchmarks.

\textbf{Methods.}
Because temporally coherent action sequences are particularly important in these robotic manipulation tasks, we compare against strong model-free baselines that utilizes action chunking. These include action-chunked flow-matching BC~\citep{flow_lipman2023}, QC~\citep{qc_li2025}, which learns an action-chunked value function and performs rejection sampling over a flow-based BC policy, and QC-IFQL, an action-chunked variant of IFQL~\citep{idql_hansenestruch2023}. We also compare with model-based baselines,  including LOMPO~\citep{rafailov2021offline} and OfflineDV2~\citep{vd4rl_lu2023} (denoted as OffDV2).

\subsubsection{Results}

Table~\ref{table:robotics} summarizes the results on visual robotic manipulation. On Robomimic, \methodname matches or exceeds BC on all three tasks and improves performance on the challenging \tt{square} task. Notably, \methodname is the only RL method that consistently matches or outperforms BC across the full benchmark. On LIBERO-10, \methodname remains competitive with BC and strong action-chunked model-free methods. We found that LIBERO-10 consists largely of unimodal demonstration data, leaving relatively limited room for rejection sampling to improve the policy over the behavioral policy. These results suggest that \methodname can retain the robustness of behavior cloning in robotic settings, while providing meaningful improvements when value-based policy selection has sufficient room to improve upon the dataset policy.

\subsection{Analysis}
\label{sec:exp_qna}

In this section, we discuss and analyze the components of \methodname through the following Q\&As. Specifically, we examine whether long-horizon model-based value expansion drives its performance gains, and which components make such expansion effective.

\clearpage
\ul{\textbf{Q: Is long-horizon model-based value expansion important for MAC-v2?}}

\begin{wrapfigure}[9]{R}{0.54\textwidth}
    \vspace{-0.8em}
    \centering
    \captionsetup{font=footnotesize, skip=3pt}    

    \begin{minipage}[t]{0.48\linewidth}
        \vspace{0pt}
        \centering
        \includegraphics[width=0.97\linewidth]{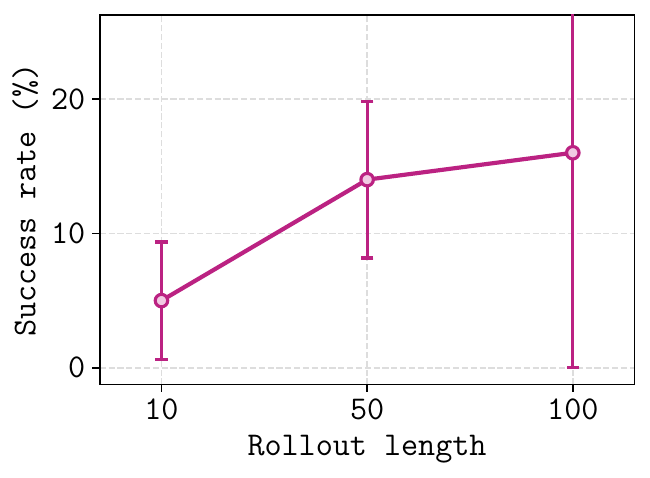}
    \end{minipage}\hfill
    \begin{minipage}[t]{0.49\linewidth}
        \vspace{0pt}
        \centering
        \includegraphics[width=0.97\linewidth]{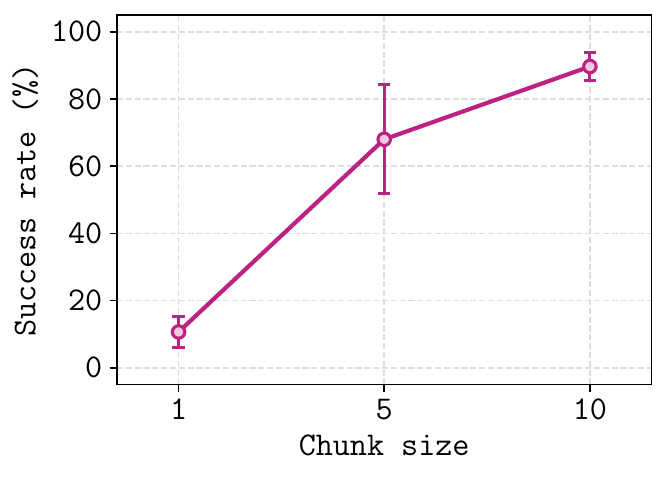}
    \end{minipage}
    \caption{\textbf{Importance of long-horizon rollout in MAC-v2.}}
    \label{fig:rollout}
\end{wrapfigure}

\textbf{A:}
To answer this question, we compare the performance of \methodname with various horizons, by varying the rollout horizon $H$ and action chunk length $n$, summarized in \Cref{fig:rollout}. First, we evaluate rollout horizons of $\{10,50,100\}$ environment steps with an action chunk size of $10$ on \tt{puzzle-4x4-task5}. Performance improves as the horizon increases, demonstrating the benefit of long-horizon value expansion. 
We also vary the action chunk size over $\{1,5,10\}$ on \tt{puzzle-4x4-task1}, and the performance improves as the chunk length increases, consistent with the result with the rollout length.
Together, these results support that long-horizon model-based value expansion with action chunking is important for the performance of \methodname.

\ul{\textbf{Q: Can prior methods also perform long-horizon value estimation?}}

\begin{wrapfigure}{r}{0.34\textwidth}
    \vspace{-1.5em}
    \centering
    \captionsetup{font=footnotesize, skip=3pt}    
    \begin{minipage}[t]{1.0\linewidth}
        \vspace{0pt}
        \centering
        \includegraphics[width=0.97\linewidth]{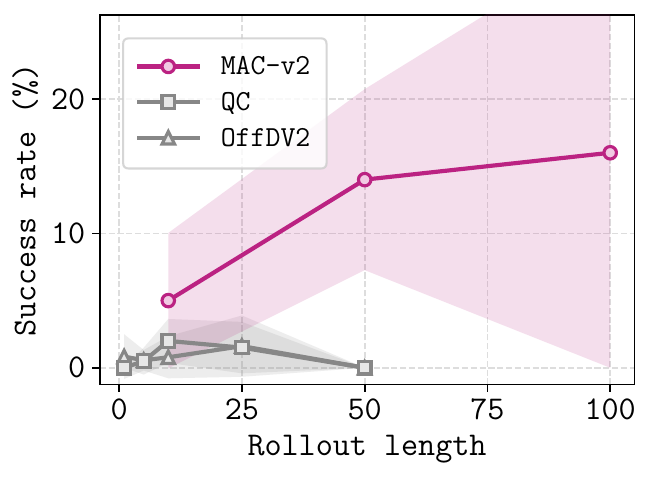}
    \end{minipage}
    \caption{\textbf{Prior works do not scale with longer horizons.}}
    \label{fig:rollout_prior}
    \vspace{-1.5em}
\end{wrapfigure}

\textbf{A:}
To answer this question, we evaluate QC~\citep{qc_li2025} and OfflineDV2~\citep{vd4rl_lu2023} on \tt{puzzle-4x4-task5} with different value estimation horizons. Specifically, we vary the action chunk size in QC and the model rollout length in OfflineDV2 to control the horizon of its value estimation. While \methodname benefits from extending the value estimation horizon, neither baseline shows comparable improvements as its horizon increases (\Cref{fig:rollout_prior}). These results suggest that simply extending the value estimation horizon is insufficient for the baselines, highlighting the importance of \methodname's design choices in facilitating long-horizon model-based value expansion.

\ul{\textbf{Q: Is each MAC-v2's objective essential for long-horizon model rollouts?}}

\begin{wrapfigure}{r}{0.35\textwidth}
    \vspace{-0.1em}
    \centering
    \raisebox{0pt}[\dimexpr\height-1.0\baselineskip\relax]{
        \begin{subfigure}[t]{1.0\linewidth}
        \includegraphics[width=\linewidth]{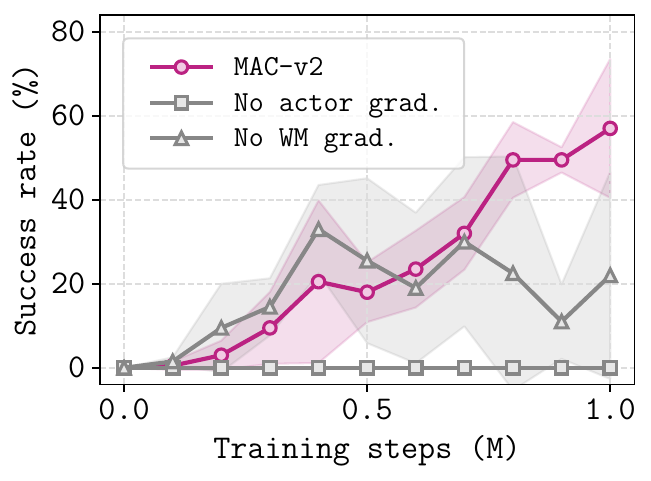}
        \end{subfigure}
    }
    \vspace{-2.0em}
    \caption{
    \footnotesize
        \textbf{Objectives are essential.}
    }
    \vspace{-1.5em}
    \label{fig:abl_actor}
\end{wrapfigure}
\textbf{A:}
We find that the both BC policy objective and WM objective plays an important role in retaining control-relevant and predictable information. Specifically, we remove the gradient from the BC objective and the world modeling objective to the encoder while keeping all other components unchanged on \tt{cube-double-task3}. \Cref{fig:abl_actor} shows that removing the actor gradient or the world modeling gradient substantially degrades performance, suggesting that both dynamics prediction and policy objective alone are insufficient to learn a representation suitable for long-horizon latent rollouts. This highlights that 
\methodname's design choice to incorporate policy objective and world modeling objective for representation learning is crucial for reliable long-horizon latent value expansion.

\ul{\textbf{Q: How about using other representation objectives instead?}}

\begin{wrapfigure}{r}{0.35\textwidth}
    \vspace{-0.1em}
    \centering
    \raisebox{0pt}[\dimexpr\height-1.0\baselineskip\relax]{
        \begin{subfigure}[t]{1.0\linewidth}
        \includegraphics[width=\linewidth]{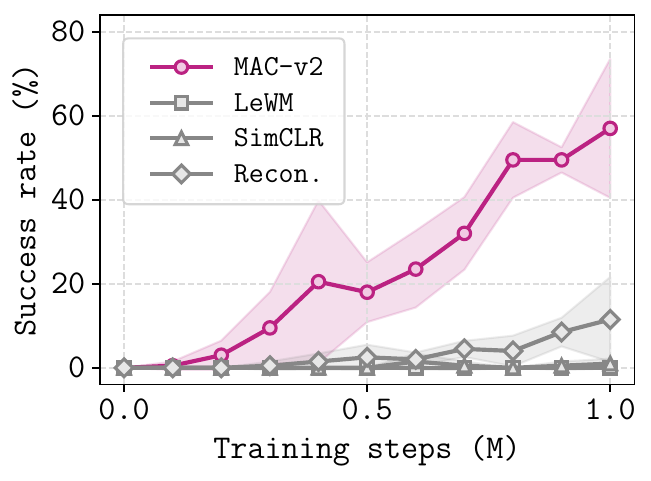}
        \end{subfigure}
    }
    \vspace{-2.0em}
    \caption{
    \footnotesize
        \textbf{Other losses falls short.}
    }
    \vspace{-2.5em}
    \label{fig:abl_recon}
\end{wrapfigure}
\textbf{A:}
Here, we replace the encoder learning objective of \methodname with other representation objectives and examine whether the learned representation can support long-horizon value expansion. We consider pixel reconstruction~\citep{wahlstrom2015pixels}, SimCLR~\citep{simclr_chen2020}, LeWorldModel~\citep{maes2026leworldmodel} and compare those methods in \tt{cube-double-task3}. 
As shown in \Cref{fig:abl_recon}, prior representation learning methods substantially degrades the performance.
This highlights that \methodname's objective is crucial for reliable long-horizon latent value expansion, and cannot be replaced with conventional representation learning objectives.

\section{Closing remarks}

In this work, we introduced \methodname, an offline model-based RL framework for long-horizon model-based value expansion from visual observations. \methodname combines reconstruction-free latent dynamics with action chunking, enabling long-horizon imagination with substantially fewer recursive world model predictions. Across 35 standard visual offline RL tasks and 13 robotic manipulation tasks, \methodname achieves strong performance against both model-free and model-based baselines, with substantial gains on challenging OGBench and Robomimic tasks while remaining competitive with strong behavior-cloning baselines on LIBERO-10. 

Several open questions remain for \methodname, suggesting directions for future work. First, while having larger action chunks reduces the number of recursive model predictions, but excessively long chunks reduce the policy's reactivity. Developing adaptive or hierarchical action chunking could provide a better trade-off between long-horizon prediction and closed-loop control. 
Moreover, our experiments are primarily limited to simulated, single-task environments with relatively limited visual variation, leaving a substantial gap to realistic multi-task settings. Extending \methodname to large-scale vision-language-action models (VLAs) or world action models (WAMs) could be a promising direction toward long-horizon model-based value expansion in more diverse and realistic domains.

\subsection*{AI use statement}

In this work, we used generative AI tools for polishing the writing, and implementing the methods. We have not used generative AI tools for designing or providing feedback on research methodology or experiments, proposing or refining hypotheses, assisting with translation. Supporting qualitative and thematic data analysis and interpreting the results, and generating synthetic data sets, helping develop theoretical models or conceptual frameworks, formulating mathematical claims, providing critical ingredients for proving mathematical claims, assisting in the writing of proofs, cleaning and reformatting dataset, interpreting results, are not applicable to this work. The authors have manually reviewed the paper to prevent plagiarism, and read the LLM-generated code and verified and tested its correctness. We take responsibility for the final content of this work, including text, claims or artifacts produced with the aid of generative AI.

\subsection*{Reproducibility statement} 

We include implementation details needed for reproduction, such as hyperparameters (including the baselines), datasets, and evaluation protocols in \Cref{sec:exp_details}. We provide the full results of each tasks in \Cref{sec:number_results} for better future reproducibility.

\subsection*{Acknowledgments}
Kwanyoung Park is partly supported by the Korea Foundation for Advanced Studies (KFAS). This research used the Savio computational cluster resource provided by the Berkeley Research Computing program at UC Berkeley. This research was partly supported by AFOSR under FA9550-22-1-0273 and DARPA TIAMAT.

\bibliography{iclr2027_conference}

\begin{thebibliography}{61}
\providecommand{\natexlab}[1]{#1}
\providecommand{\url}[1]{\texttt{#1}}
\expandafter\ifx\csname urlstyle\endcsname\relax
  \providecommand{\doi}[1]{doi: #1}\else
  \providecommand{\doi}{doi: \begingroup \urlstyle{rm}\Url}\fi

\bibitem[Albergo \& Vanden-Eijnden(2023)Albergo and Vanden-Eijnden]{flow_albergo2023}
Michael~S Albergo and Eric Vanden-Eijnden.
\newblock Building normalizing flows with stochastic interpolants.
\newblock In \emph{International Conference on Learning Representations (ICLR)}, 2023.

\bibitem[Chen et~al.(2023)Chen, Lu, Ying, Su, and Zhu]{sfbc_chen2023}
Huayu Chen, Cheng Lu, Chengyang Ying, Hang Su, and Jun Zhu.
\newblock Offline reinforcement learning via high-fidelity generative behavior modeling.
\newblock In \emph{International Conference on Learning Representations (ICLR)}, 2023.

\bibitem[Chen et~al.(2020)Chen, Kornblith, Norouzi, and Hinton]{simclr_chen2020}
Ting Chen, Simon Kornblith, Mohammad Norouzi, and Geoffrey~E. Hinton.
\newblock A simple framework for contrastive learning of visual representations.
\newblock In \emph{International Conference on Machine Learning (ICML)}, 2020.

\bibitem[Chung et~al.(2026)Chung, Lee, and Oh]{chung2026offline}
Hojun Chung, Junseo Lee, and Songhwai Oh.
\newblock Offline reinforcement learning with universal horizon models.
\newblock \emph{arXiv preprint arXiv:2605.15603}, 2026.

\bibitem[Espeholt et~al.(2018)Espeholt, Soyer, Munos, Simonyan, Mnih, Ward, Doron, Firoiu, Harley, Dunning, Legg, and Kavukcuoglu]{impala_espeholt2018}
Lasse Espeholt, Hubert Soyer, R{\'e}mi Munos, Karen Simonyan, Volodymyr Mnih, Tom Ward, Yotam Doron, Vlad Firoiu, Tim Harley, Iain Dunning, Shane Legg, and Koray Kavukcuoglu.
\newblock Impala: Scalable distributed deep-rl with importance weighted actor-learner architectures.
\newblock In \emph{International Conference on Machine Learning (ICML)}, 2018.

\bibitem[Feinberg et~al.(2018)Feinberg, Wan, Stoica, Jordan, Gonzalez, and Levine]{mve_feinberg2018}
Vladimir Feinberg, Alvin Wan, Ion Stoica, Michael~I Jordan, Joseph~E Gonzalez, and Sergey Levine.
\newblock Model-based value expansion for efficient model-free reinforcement learning.
\newblock In \emph{International Conference on Machine Learning (ICML)}, 2018.

\bibitem[Fu et~al.(2026)Fu, Feng, Hansen, and Huang]{fu2026back}
Minghao Fu, Fan Feng, Nicklas Hansen, and Biwei Huang.
\newblock Back to parsimonious latents: Learning task-centric world models from visual foundations.
\newblock \emph{arXiv preprint arXiv:2605.25620}, 2026.

\bibitem[Fujimoto \& Gu(2021)Fujimoto and Gu]{td3bc_fujimoto2021}
Scott Fujimoto and Shixiang~Shane Gu.
\newblock A minimalist approach to offline reinforcement learning.
\newblock In \emph{Neural Information Processing Systems (NeurIPS)}, 2021.

\bibitem[Garg et~al.(2023)Garg, Hejna, Geist, and Ermon]{xql_garg2023}
Divyansh Garg, Joey Hejna, Matthieu Geist, and Stefano Ermon.
\newblock Extreme q-learning: Maxent rl without entropy.
\newblock In \emph{International Conference on Learning Representations (ICLR)}, 2023.

\bibitem[Hafner et~al.(2020{\natexlab{a}})Hafner, Lillicrap, Norouzi, and Ba]{hafner2020dreamerv2}
Danijar Hafner, Timothy Lillicrap, Mohammad Norouzi, and Jimmy Ba.
\newblock Mastering atari with discrete world models.
\newblock \emph{arXiv preprint arXiv:2010.02193}, 2020{\natexlab{a}}.

\bibitem[Hafner et~al.(2020{\natexlab{b}})Hafner, Lillicrap, Ba, and Norouzi]{dreamer_hafner2020}
Danijar Hafner, Timothy~P. Lillicrap, Jimmy Ba, and Mohammad Norouzi.
\newblock Dream to control: Learning behaviors by latent imagination.
\newblock In \emph{International Conference on Learning Representations (ICLR)}, 2020{\natexlab{b}}.

\bibitem[Hafner et~al.(2025{\natexlab{a}})Hafner, Pasukonis, Ba, and Lillicrap]{dreamerv3_hafner2025}
Danijar Hafner, Jurgis Pasukonis, Jimmy Ba, and Timothy Lillicrap.
\newblock Mastering diverse control tasks through world models.
\newblock \emph{Nature}, 640:\penalty0 647--653, 2025{\natexlab{a}}.

\bibitem[Hafner et~al.(2025{\natexlab{b}})Hafner, Yan, and Lillicrap]{dreamerv4_hafner2025}
Danijar Hafner, Wilson Yan, and Timothy Lillicrap.
\newblock Training agents inside of scalable world models.
\newblock \emph{arXiv preprint arXiv:2509.24527}, 2025{\natexlab{b}}.

\bibitem[Hansen \& Wang(2026)Hansen and Wang]{Hansen2026Hallucination}
Nicklas Hansen and Xiaolong Wang.
\newblock Hallucination in world models is predictable and preventable, 2026.

\bibitem[Hansen et~al.(2022)Hansen, Wang, and Su]{tdmpc_hansen2022}
Nicklas Hansen, Xiaolong Wang, and Hao Su.
\newblock Temporal difference learning for model predictive control.
\newblock In \emph{International Conference on Machine Learning (ICML)}, 2022.

\bibitem[Hansen et~al.(2024)Hansen, Su, and Wang]{tdmpc2_hansen2024}
Nicklas Hansen, Hao Su, and Xiaolong Wang.
\newblock Td-mpc2: Scalable, robust world models for continuous control.
\newblock In \emph{International Conference on Learning Representations (ICLR)}, 2024.

\bibitem[Hansen-Estruch et~al.(2023)Hansen-Estruch, Kostrikov, Janner, Kuba, and Levine]{idql_hansenestruch2023}
Philippe Hansen-Estruch, Ilya Kostrikov, Michael Janner, Jakub~Grudzien Kuba, and Sergey Levine.
\newblock Idql: Implicit q-learning as an actor-critic method with diffusion policies.
\newblock \emph{ArXiv}, abs/2304.10573, 2023.

\bibitem[Islam et~al.(2022)Islam, Tomar, Lamb, Efroni, Zang, Didolkar, Misra, Li, Van~Seijen, Combes, et~al.]{islam2022agent}
Riashat Islam, Manan Tomar, Alex Lamb, Yonathan Efroni, Hongyu Zang, Aniket Didolkar, Dipendra Misra, Xin Li, Harm Van~Seijen, Remi Tachet~des Combes, et~al.
\newblock Agent-controller representations: Principled offline rl with rich exogenous information.
\newblock \emph{arXiv preprint arXiv:2211.00164}, 2022.

\bibitem[Kidambi et~al.(2020)Kidambi, Rajeswaran, Netrapalli, and Joachims]{morel_kidambi2020}
Rahul Kidambi, Aravind Rajeswaran, Praneeth Netrapalli, and Thorsten Joachims.
\newblock Morel : Model-based offline reinforcement learning.
\newblock In \emph{Neural Information Processing Systems (NeurIPS)}, 2020.

\bibitem[Kingma \& Ba(2015)Kingma and Ba]{adam_kingma2015}
Diederik~P. Kingma and Jimmy Ba.
\newblock Adam: A method for stochastic optimization.
\newblock In \emph{International Conference on Learning Representations (ICLR)}, 2015.

\bibitem[Kostrikov et~al.(2021)Kostrikov, Yarats, and Fergus]{drq_kostrikov2021}
Ilya Kostrikov, Denis Yarats, and Rob Fergus.
\newblock Image augmentation is all you need: Regularizing deep reinforcement learning from pixels.
\newblock In \emph{International Conference on Learning Representations (ICLR)}, 2021.

\bibitem[Kostrikov et~al.(2022)Kostrikov, Nair, and Levine]{iql_kostrikov2022}
Ilya Kostrikov, Ashvin Nair, and Sergey Levine.
\newblock Offline reinforcement learning with implicit q-learning.
\newblock In \emph{International Conference on Learning Representations (ICLR)}, 2022.

\bibitem[Kumar et~al.(2020)Kumar, Zhou, Tucker, and Levine]{cql_kumar2020}
Aviral Kumar, Aurick Zhou, G.~Tucker, and Sergey Levine.
\newblock Conservative q-learning for offline reinforcement learning.
\newblock In \emph{Neural Information Processing Systems (NeurIPS)}, 2020.

\bibitem[Lange et~al.(2012)Lange, Gabel, and Riedmiller]{batch_lange2012}
Sascha Lange, Thomas Gabel, and Martin Riedmiller.
\newblock Batch reinforcement learning.
\newblock In \emph{Reinforcement learning: State-of-the-art}, pp.\  45--73. Springer, 2012.

\bibitem[Laskin et~al.(2020)Laskin, Srinivas, and Abbeel]{laskin2020curl}
Michael Laskin, Aravind Srinivas, and Pieter Abbeel.
\newblock Curl: Contrastive unsupervised representations for reinforcement learning.
\newblock In \emph{International conference on machine learning}, pp.\  5639--5650. PMLR, 2020.

\bibitem[Lee et~al.(2025)Lee, Hwang, Kim, Kim, Tai, Subramanian, Wurman, Choo, Stone, and Seno]{simba_lee2025}
Hojoon Lee, Dongyoon Hwang, Donghu Kim, Hyunseung Kim, Jun~Jet Tai, Kaushik Subramanian, Peter~R Wurman, Jaegul Choo, Peter Stone, and Takuma Seno.
\newblock Simba: Simplicity bias for scaling up parameters in deep reinforcement learning.
\newblock In \emph{International Conference on Learning Representations (ICLR)}, 2025.

\bibitem[Lee(2012)]{smooth_lee2002}
John~M Lee.
\newblock \emph{Introduction to Smooth Manifolds}.
\newblock Springer, 2012.

\bibitem[Levine et~al.(2020)Levine, Kumar, Tucker, and Fu]{offline_levine2020}
Sergey Levine, Aviral Kumar, G.~Tucker, and Justin Fu.
\newblock Offline reinforcement learning: Tutorial, review, and perspectives on open problems.
\newblock \emph{ArXiv}, abs/2005.01643, 2020.

\bibitem[Li et~al.(2025{\natexlab{a}})Li, Park, and Levine]{dqc_li2025}
Qiyang Li, Seohong Park, and Sergey Levine.
\newblock Decoupled q-chunking.
\newblock \emph{arXiv preprint arXiv:2512.10926}, 2025{\natexlab{a}}.

\bibitem[Li et~al.(2025{\natexlab{b}})Li, Zhou, and Levine]{qc_li2025}
Qiyang Li, Zhiyuan Zhou, and Sergey Levine.
\newblock Reinforcement learning with action chunking.
\newblock \emph{arXiv preprint arXiv:2507.07969}, 2025{\natexlab{b}}.

\bibitem[Lipman et~al.(2023)Lipman, Chen, Ben-Hamu, Nickel, and Le]{flow_lipman2023}
Yaron Lipman, Ricky~TQ Chen, Heli Ben-Hamu, Maximilian Nickel, and Matt Le.
\newblock Flow matching for generative modeling.
\newblock In \emph{International Conference on Learning Representations (ICLR)}, 2023.

\bibitem[Liu et~al.(2023{\natexlab{a}})Liu, Zhu, Gao, Feng, Liu, Zhu, and Stone]{liu2023libero}
Bo~Liu, Yifeng Zhu, Chongkai Gao, Yihao Feng, Qiang Liu, Yuke Zhu, and Peter Stone.
\newblock Libero: Benchmarking knowledge transfer for lifelong robot learning.
\newblock \emph{arXiv preprint arXiv:2306.03310}, 2023{\natexlab{a}}.

\bibitem[Liu et~al.(2023{\natexlab{b}})Liu, Gong, and Liu]{flow_liu2023}
Xingchao Liu, Chengyue Gong, and Qiang Liu.
\newblock Flow straight and fast: Learning to generate and transfer data with rectified flow.
\newblock In \emph{International Conference on Learning Representations (ICLR)}, 2023{\natexlab{b}}.

\bibitem[Lu et~al.(2023)Lu, Ball, Rudner, Parker-Holder, Osborne, and Teh]{vd4rl_lu2023}
Cong Lu, Philip~J Ball, Tim~GJ Rudner, Jack Parker-Holder, Michael~A Osborne, and Yee~Whye Teh.
\newblock Challenges and opportunities in offline reinforcement learning from visual observations.
\newblock \emph{Transactions on Machine Learning Research (TMLR)}, 2023.

\bibitem[Luo et~al.(2026)Luo, Zhang, Feng, Zheng, Xu, Xu, Xi, Fu, and Lu]{luo2026being}
Hao Luo, Wanpeng Zhang, Yicheng Feng, Sipeng Zheng, Haiweng Xu, Chaoyi Xu, Ziheng Xi, Yuhui Fu, and Zongqing Lu.
\newblock Being-h0. 7: A latent world-action model from egocentric videos.
\newblock \emph{arXiv preprint arXiv:2605.00078}, 2026.

\bibitem[Maes et~al.(2026)Maes, Lidec, Scieur, LeCun, and Balestriero]{maes2026leworldmodel}
Lucas Maes, Quentin~Le Lidec, Damien Scieur, Yann LeCun, and Randall Balestriero.
\newblock Leworldmodel: Stable end-to-end joint-embedding predictive architecture from pixels.
\newblock \emph{arXiv preprint arXiv:2603.19312}, 2026.

\bibitem[Mandlekar et~al.(2021)Mandlekar, Xu, Wong, Nasiriany, Wang, Kulkarni, Fei-Fei, Savarese, Zhu, and Mart\'{i}n-Mart\'{i}n]{robomimic_mandlekar2021}
Ajay Mandlekar, Danfei Xu, J.~Wong, Soroush Nasiriany, Chen Wang, Rohun Kulkarni, Li~Fei-Fei, Silvio Savarese, Yuke Zhu, and Roberto Mart\'{i}n-Mart\'{i}n.
\newblock What matters in learning from offline human demonstrations for robot manipulation.
\newblock In \emph{Conference on Robot Learning (CoRL)}, 2021.

\bibitem[Mnih et~al.(2013)Mnih, Kavukcuoglu, Silver, Graves, Antonoglou, Wierstra, and Riedmiller]{dqn_mnih2013}
Volodymyr Mnih, Koray Kavukcuoglu, David Silver, Alex Graves, Ioannis Antonoglou, Daan Wierstra, and Martin~A. Riedmiller.
\newblock Playing atari with deep reinforcement learning.
\newblock \emph{ArXiv}, abs/1312.5602, 2013.

\bibitem[Nair et~al.(2020)Nair, Dalal, Gupta, and Levine]{awac_nair2020}
Ashvin Nair, Murtaza Dalal, Abhishek Gupta, and Sergey Levine.
\newblock Accelerating online reinforcement learning with offline datasets.
\newblock \emph{ArXiv}, abs/2006.09359, 2020.

\bibitem[Park \& Lee(2025)Park and Lee]{leq_park2025}
Kwanyoung Park and Youngwoon Lee.
\newblock Model-based offline reinforcement learning with lower expectile q-learning.
\newblock \emph{International Conference on Learning Representations (ICLR)}, 2025.

\bibitem[Park et~al.(2026)Park, Park, Lee, and Levine]{mac_park2025}
Kwanyoung Park, Seohong Park, Youngwoon Lee, and Sergey Levine.
\newblock Scalable offline model-based rl with action chunks.
\newblock \emph{International Conference on Learning Representations (ICLR)}, 2026.

\bibitem[Park et~al.(2025{\natexlab{a}})Park, Frans, Eysenbach, and Levine]{ogbench_park2025}
Seohong Park, Kevin Frans, Benjamin Eysenbach, and Sergey Levine.
\newblock Ogbench: Benchmarking offline goal-conditioned rl.
\newblock In \emph{International Conference on Learning Representations (ICLR)}, 2025{\natexlab{a}}.

\bibitem[Park et~al.(2025{\natexlab{b}})Park, Frans, Mann, Eysenbach, Kumar, and Levine]{sharsa_park2025}
Seohong Park, Kevin Frans, Deepinder Mann, Benjamin Eysenbach, Aviral Kumar, and Sergey Levine.
\newblock Horizon reduction makes rl scalable.
\newblock \emph{arXiv}, abs/2506.04168, 2025{\natexlab{b}}.

\bibitem[Park et~al.(2025{\natexlab{c}})Park, Li, and Levine]{fql_park2025}
Seohong Park, Qiyang Li, and Sergey Levine.
\newblock Flow q-learning.
\newblock In \emph{International Conference on Machine Learning (ICML)}, 2025{\natexlab{c}}.

\bibitem[Peng et~al.(2019)Peng, Kumar, Zhang, and Levine]{awr_peng2019}
Xue~Bin Peng, Aviral Kumar, Grace Zhang, and Sergey Levine.
\newblock Advantage-weighted regression: Simple and scalable off-policy reinforcement learning.
\newblock \emph{ArXiv}, abs/1910.00177, 2019.

\bibitem[Rafailov et~al.(2021)Rafailov, Yu, Rajeswaran, and Finn]{rafailov2021offline}
Rafael Rafailov, Tianhe Yu, Aravind Rajeswaran, and Chelsea Finn.
\newblock Offline reinforcement learning from images with latent space models.
\newblock In \emph{Learning for dynamics and control}, pp.\  1154--1168. PMLR, 2021.

\bibitem[Rigter et~al.(2022)Rigter, Lacerda, and Hawes]{rambo_rigter2022}
Marc Rigter, Bruno Lacerda, and Nick Hawes.
\newblock Rambo-rl: Robust adversarial model-based offline reinforcement learning.
\newblock \emph{Neural Information Processing Systems (NeurIPS)}, 2022.

\bibitem[Schrittwieser et~al.(2020)Schrittwieser, Antonoglou, Hubert, Simonyan, Sifre, Schmitt, Guez, Lockhart, Hassabis, Graepel, Lillicrap, and Silver]{muzero_schrittwieser2020}
Julian Schrittwieser, Ioannis Antonoglou, T.~Hubert, K.~Simonyan, L.~Sifre, Simon Schmitt, A.~Guez, Edward Lockhart, Demis Hassabis, T.~Graepel, T.~Lillicrap, and D.~Silver.
\newblock Mastering {Atari, Go, Chess and Shogi} by planning with a learned model.
\newblock \emph{Nature}, 588 7839:\penalty0 604--609, 2020.

\bibitem[Sims et~al.(2024)Sims, Lu, Foerster, and Teh]{ravl_sims2024}
Anya Sims, Cong Lu, Jakob~N Foerster, and Yee~W Teh.
\newblock The edge-of-reach problem in offline model-based reinforcement learning.
\newblock \emph{Advances in Neural Information Processing Systems (NeurIPS)}, 2024.

\bibitem[Song et~al.(2026)Song, Park, and Lee]{cgql_song2026}
Gwanwoo Song, Kwanyoung Park, and Youngwoon Lee.
\newblock Chunk-guided q-learning.
\newblock \emph{arXiv preprint arXiv:2603.13971}, 2026.

\bibitem[Stooke et~al.(2021)Stooke, Lee, Abbeel, and Laskin]{stooke2021decoupling}
Adam Stooke, Kimin Lee, Pieter Abbeel, and Michael Laskin.
\newblock Decoupling representation learning from reinforcement learning.
\newblock In \emph{International conference on machine learning}, pp.\  9870--9879. PMLR, 2021.

\bibitem[Sun et~al.(2024)Sun, Zang, Li, and Islam]{sun2024learning}
Ruixiang Sun, Hongyu Zang, Xin Li, and Riashat Islam.
\newblock Learning latent dynamic robust representations for world models.
\newblock \emph{arXiv preprint arXiv:2405.06263}, 2024.

\bibitem[Sun et~al.(2023)Sun, Zhang, Jia, Lin, Ye, and Yu]{mobile_sun2023}
Yihao Sun, Jiaji Zhang, Chengxing Jia, Haoxin Lin, Junyin Ye, and Yang Yu.
\newblock Model-bellman inconsistency for model-based offline reinforcement learning.
\newblock In \emph{International Conference on Machine Learning}, pp.\  33177--33194. PMLR, 2023.

\bibitem[Tarasov et~al.(2023)Tarasov, Kurenkov, Nikulin, and Kolesnikov]{rebrac_tarasov2023}
Denis Tarasov, Vladislav Kurenkov, Alexander Nikulin, and Sergey Kolesnikov.
\newblock Revisiting the minimalist approach to offline reinforcement learning.
\newblock In \emph{Neural Information Processing Systems (NeurIPS)}, 2023.

\bibitem[Wahlstr{\"o}m et~al.(2015)Wahlstr{\"o}m, Sch{\"o}n, and Deisenroth]{wahlstrom2015pixels}
Niklas Wahlstr{\"o}m, Thomas~B Sch{\"o}n, and Marc~Peter Deisenroth.
\newblock From pixels to torques: Policy learning with deep dynamical models.
\newblock \emph{arXiv preprint arXiv:1502.02251}, 2015.

\bibitem[Wan et~al.(2024)Wan, Chen, Gan, Feng, and Zhan]{wan2024semopo}
Shenghua Wan, Ziyuan Chen, Le~Gan, Shuai Feng, and De-Chuan Zhan.
\newblock Semopo: learning high-quality model and policy from low-quality offline visual datasets.
\newblock In \emph{Proceedings of the 41st International Conference on Machine Learning}, pp.\  49867--49889, 2024.

\bibitem[Wu et~al.(2019)Wu, Tucker, and Nachum]{brac_wu2019}
Yifan Wu, G.~Tucker, and Ofir Nachum.
\newblock Behavior regularized offline reinforcement learning.
\newblock \emph{ArXiv}, abs/1911.11361, 2019.

\bibitem[Xu et~al.(2023)Xu, Jiang, Li, Yang, Wang, Chan, and Zhan]{sql_xu2023}
Haoran Xu, Li~Jiang, Jianxiong Li, Zhuoran Yang, Zhaoran Wang, Victor Chan, and Xianyuan Zhan.
\newblock Offline rl with no ood actions: In-sample learning via implicit value regularization.
\newblock In \emph{International Conference on Learning Representations (ICLR)}, 2023.

\bibitem[Yu et~al.(2020)Yu, Thomas, Yu, Ermon, Zou, Levine, Finn, and Ma]{mopo_yu2020}
Tianhe Yu, Garrett Thomas, Lantao Yu, Stefano Ermon, James~Y. Zou, Sergey Levine, Chelsea Finn, and Tengyu Ma.
\newblock Mopo: Model-based offline policy optimization.
\newblock In \emph{Neural Information Processing Systems (NeurIPS)}, 2020.

\bibitem[Yu et~al.(2021)Yu, Kumar, Rafailov, Rajeswaran, Levine, and Finn]{combo_yu2021}
Tianhe Yu, Aviral Kumar, Rafael Rafailov, Aravind Rajeswaran, Sergey Levine, and Chelsea Finn.
\newblock Combo: Conservative offline model-based policy optimization.
\newblock In \emph{Neural Information Processing Systems (NeurIPS)}, 2021.

\bibitem[Zhao et~al.(2023)Zhao, Kumar, Levine, and Finn]{aloha_zhao2023}
Tony~Z Zhao, Vikash Kumar, Sergey Levine, and Chelsea Finn.
\newblock Learning fine-grained bimanual manipulation with low-cost hardware.
\newblock In \emph{Robotics: Science and Systems (RSS)}, 2023.

\end{thebibliography}
\bibliographystyle{iclr2027_conference}

\clearpage
\appendix

\section{Experimental Details}
\label{sec:exp_details}

We implement \methodname on top of the codebase of~\citet{sharsa_park2025}. Each experiment takes approximately 9 hours for OGBench and MMBench2 benchmarks, 12 hours for Robomimic and 14 hours for LIBERO-10 benchmark on a single A5000 GPU.

\subsection{Implementation Details}

\label{subsec:details}
\textbf{Network architecture for \methodname.} 
For OGBench experiments, we use the \tt{impala\_small} encoder from~\citet{impala_espeholt2018}, following the implementation of FQL~\citep{fql_park2025} to match the baselines.
For LIBERO-10 experiments, we use the Resnet-18 backbone (trained from scratch) as the image is more complicated and higher resolution (128px). For other experiments, we use the convolutional encoder architecture (12M version) from DreamerV3~\citep{dreamerv3_hafner2025}. On top of the encoder, we use 4-layer MLP with hidden size 256 for actor, critic, and dynamics head. For critic and reward head, we use the two-hot encoding~\citep{dreamerv3_hafner2025}. 

\textbf{Data augmentation.} We apply random crop augmentation of size $4$ for OGBench and $12$ for LIBERO-10, following the implementation of~\citet{fql_park2025}. For model-free approaches, we use the same augmentation scheme as they follow the implementation of~\citet{fql_park2025}. For model-based RL baselines, we follow the setup of original papers.

\textbf{Accelerating rejection sampling during training.}
Following~\citet{mac_park2025}, we use different numbers of samples for rejection sampling during training and evaluation (which we call $N_\mathrm{train}$ and $N_\mathrm{test}$) to reduce the training time.
Specifically,
we use $N_{\mathrm{train}}=8$ during training 
and $N_{\mathrm{test}}=32$ at test time. 

\textbf{Implementation details for the compared methods.} 
For the model-free baselines, we follow the implementation by~\citet{ogbench_park2025} and~\citet{qc_li2025}. For LOMPO~\citep{rafailov2021offline} and OfflineDV2~\citep{vd4rl_lu2023}, we use the official codebase released by the authors.

\textbf{\methodname with terminals.} We predict the terminal signal $\mathbf{\hat{d}}_{t} = 1 - \prod_{i=0}^{n-1} (1-\hat d_{t+i})$ from the terminal model $d_{\psi}(z_t, \bm{a}_t): \mathcal{Z} \times \mathcal{A}^\achunk \rightarrow \mathbb{R}$, and calculate the discounted reward target $\bm{r}_t = \sum_{i=0}^{\achunk-1} \gamma^i (\prod_{j=1}^{i-1} (1 - d_{t+j})) r_t$. We use a hard threshold $0.5$ and calculate $\hat d_{t} = \mathbbm{1}(d_{\psi}(z_t, a_t) > 0.5)$.
We train the value function from latent on-policy model rollout (i.e., imaginary trajectories), generated as 
\begin{align}
    \mathcal{T} = \{(z_t^{(i)}, \bm{\hat{a}}_{t}^{(i)}, \bm{\hat r}_{t}^{(i)}, \bm{\hat d}_{t}^{(i)}, \hat z_{t+\achunk}^{(i)}, \bm{\hat{a}}_{t+\achunk}^{(i)}, \bm{\hat r}_{t+\achunk}^{(i)}, \bm{\hat d}_{t+\achunk}^{(i)},\ldots, \hat z_{t+H\achunk}^{(i)})\}_{i=1}^M.
\end{align}

We calculate the value target as:
\begin{align}
\hat v_{t+kn} = \sum_{i=k}^{H-1} \gamma^{(i-k)n}  (\prod_{j=k}^{i-1} \hat d_{t+jn}) \bm{\hat r}_{t+in} + \gamma^{(H-k)n}(\prod_{j=k}^{H-1} \hat d_{t+jn})v_{\bar \phi}(\hat s_{t+Hn}),
\end{align}

Finally, we train the action-chunked Q function 
$Q_\phi(\pl{s_t}, \pl{\bm{a}_{t}}): \gZ \times \gA^n \to \Delta(\sR)$ for the rejection sampling
with the following loss:
\begin{align}
L^Q(\phi) &= \E_{o_t \sim \gD} \left[-\log 
Q_\phi(\bm{\hat r}_{t} + \gamma^n (1-\bm{\hat d}_{t+j\achunk}) v_\phi(\hat z_{t+n}) \mid z_{t}, \bm{\hat a}_{t})\right]. \label{eq:loss_q_app}
\end{align}

\clearpage
\subsection{Hyperparameters}

We group the configurations into three families:
model-free baselines (\Cref{tab:hyperparameters_modelfree}), model-based baselines (\Cref{tab:hyperparameters_modelbased}), and \methodname{} (\Cref{tab:hyperparameters}). Image resolution, encoder and augmentation are set per benchmark, and are given in \Cref{tab:hyperparameters_perbench}. Evaluation uses $15$ episodes for each seed. We match the encoder for \methodname and model-free methods, but we use the original codebase's architecture for model-based baselines as their architecture (history-conditioned RSSM) significantly differs from our method (markovian CNN encoder). 

\begin{table}[H]
    \small
    \centering
    \caption{\footnotesize\textbf{Per-benchmark input and encoder settings.} These
    apply to \methodname{} and the model-free baselines, so every comparison
    uses the same encoder within a benchmark. Please refer to \Cref{subsec:details} for implementation details of the encoder.}
    \label{tab:hyperparameters_perbench}
    \begin{tabular}{llll}
    \toprule
    \textbf{Benchmark} & \textbf{Resolution} & \textbf{Encoder} & \textbf{Aug.\ pad} \\
    \midrule
    OGBench    & $64 \times 64$   & \texttt{impala\_small} & $4$  \\
    MMBench2   & $64 \times 64$   & \texttt{12M Conv.}          & $4$  \\
    Robomimic  & $84 \times 84$   & \texttt{12M Conv.}         & $4$  \\
    LIBERO-10  & $128 \times 128$ & \texttt{ResNet-18}     & $12$ \\
    \bottomrule
    \end{tabular}
\end{table}

\paragraph{Shared hyperparameters for model-free baseline.}
IQL, ReBRAC, IFQL, FQL and QC are run from the reference FQL
codebase~\citep{fql_park2025} at its visual-task defaults. QC additionally uses
best-of-$32$ sampling over $5$-step action chunks, so it is chunked where the
other four are single-step. We use $p_{\mathrm{aug}} = 0.5$ following the original codebase.

\begin{table}[H]
    \small
    \centering
    \caption{\footnotesize\textbf{Hyperparameters of the model-free baselines.}}
    \label{tab:hyperparameters_modelfree}
    \begin{tabular}{ll}
    \toprule
    \textbf{Hyperparameters} & \textbf{Value} \\
    \midrule
    Gradient steps      & $500$K \\
    Learning rate       & $3 \times 10^{-4}$ \\
    Optimizer           & Adam~\citep{adam_kingma2015} \\
    Batch size          & $256$ \\
    MLP size            & $[512, 512, 512, 512]$ \\
    Frame stack         & $3$ \\
    Image augmentation  & DrQ random shift, $p_{\mathrm{aug}}=0.5$ (tied) \\
    Discount factor $\gamma$ & $0.99$ \\
    \bottomrule
    \end{tabular}
\end{table}

\paragraph{Shared hyperparameters for model-based baseline.}
Both are run at their authors' defaults, changing only the uncertainty
coefficient $\lambda$. 
Both methods train their world model for $25$K steps before the policy phase. Both methods are trained for $100$k steps, as we found the performance starts to degrade after $100$k steps. We note that the computational budget for $100$k steps for these methods is larger than MAC-v2, taking around 16 hours in a single A5000 GPU.

\begin{table}[H]
    \small
    \centering
    \caption{\footnotesize\textbf{Shared hyperparameters of model-based baselines.}}
    \label{tab:hyperparameters_modelbased}
    \begin{tabular}{ll}
    \toprule
    \textbf{Hyperparameters} & \textbf{Value} \\
    \midrule
    Gradient steps  & $100$K  \\
    Learning rate   & $3 \times 10^{-4}$  \\
    Optimizer       & Adam~\citep{adam_kingma2015}  \\
    Batch size & $256$  \\
    MLP size & $[512, 512, 512, 512]$ \\
    Image resolution & $64 \times 64$ \\
    Frame stack & $1$ \\
    \bottomrule
    \end{tabular}
\end{table}

\paragraph{\methodname hyperparameters.}
We report the hyperparameters for our method in \Cref{tab:hyperparameters}.
\methodname uses the same
$(\achunk, H, N_\mathrm{train}, N_\mathrm{test}) = (10, 10, 8, 32)$ across all tasks and all four benchmarks, no per-task tuning is performed.
We apply DrQ-style random-shift augmentation~\citep{drq_kostrikov2021}, using the same shift for an observation and the next observation. We normalize the value with the maximum return of the dataset, and train the value and reward heads with symexp two-hot categorical loss of~\citet{dreamerv3_hafner2025}.

\begin{table}[H]
    \small
    \centering
    \caption{\footnotesize\textbf{Hyperparameters of \methodname.}}
    \label{tab:hyperparameters}
    \begin{tabular}{ll}
    \toprule
    \textbf{Hyperparameters} & \textbf{Value} \\
    \midrule
    Learning rate   & $3 \times 10^{-4}$  \\
    Optimizer       & Adam  \\
    Layer normalization & True \\
    Target network update rate $\tau$ & $0.02$ \\
    Discount factor $\gamma$ & $0.999$ \\
    $\lambda$-return coefficient & $1.0$ \\
    Flow steps & $5$ \\
    $N_{\mathrm{train}}$  & $8$ \\
    $N_{\mathrm{test}}$   & $32$ \\
    Rollout length $H$  & $10$   \\
    Action chunk size $\achunk$ & $10$  \\
    Latent dimension & $512$ \\
    Image augmentation & DrQ random shift (tied) \\
    Value / reward loss & symexp two-hot, $255$ bins on $[-5, 5]$ \\
    Weight decay & $1 \times 10^{-4}$ \\
    Gradient clipping & $1.0$ \\
    Frame stack         & $1$ \\
    \bottomrule
    \end{tabular}
\end{table}

\textbf{Hyperparameters for baselines.}
For LOMPO and Offline DV2, we sweep the uncertainty penalty coefficient
$\lambda \in \{1, 3, 10, 30\}$. For FQL we search over the behavior-cloning coefficient $\alpha \in \{0.1, 0.3, 1.0\}$, and for ReBRAC over $(\alpha_{\mathrm{actor}}, \alpha_{\mathrm{critic}}) \in \{0.01, 0.1, 1.0\} \times \{0, 0.01\}$. For FQL, IQL, ReBRAC, we use the reported hyperparameter from~\citet{fql_park2025} in OGBench. For QC and QC-IFQL, we search over the action chunking size $\{5, 10\}$. IFQL is run at a single fixed configuration. The hyperparameters are reported in Table~\ref{tab:baseline_hyperparameters_ogbench},~\ref{tab:baseline_hyperparameters_mmbench2},~\ref{tab:baseline_hyperparameters_robomimic_chunk},~\ref{tab:baseline_hyperparameters_libero10} for OGBench, MMBench2, Robomimic, LIBERO-10, repectively.
We denote ``N/A'' in the tables if a method achieves zero performance across all hyperparameters tested in our sweep. If not specified, all other hyperparameters follow the defaults provided in the original papers.

\begin{table}[h]
    \small
    \centering
    \caption{\footnotesize\textbf{Baseline hyperparameters on the
    OGBench visual tasks.} Uncertainty penalty $\lambda$ for the model-based baselines, behavior-cloning
    coefficient $\alpha$ for FQL, and $(\alpha_{\mathrm{actor}}, \alpha_{\mathrm{critic}})$
    for ReBRAC.}
    \label{tab:baseline_hyperparameters_ogbench}
    \begin{tabular}{llllll}
    \toprule
    \tt{Environment} & \tt{LOMPO} $(\lambda)$ & \tt{OffDV2} $(\lambda)$ &
    \tt{FQL} $(\alpha)$ & \tt{ReBRAC} $(\alpha_{\mathrm{actor}}, \alpha_{\mathrm{critic}})$ & \tt{IQL} $(\alpha)$ \\
    \midrule
    \texttt{visual-cube-single-play-v0} & $5$ & $5$  & $300$ & $(1.0, 0)$    & $1$  \\
    \texttt{visual-cube-double-play-v0} & $5$ & $10$ & $100$ & $(0.1, 0)$    & $0.3$ \\
    \texttt{visual-scene-play-v0}       & $3$ & $10$ & $100$ & $(0.1, 0.01)$ & $10$ \\
    \texttt{visual-puzzle-3x3-play-v0}  & $3$ & $10$ & $300$ & $(0.3, 0.01)$ & $10$ \\
    \texttt{visual-puzzle-4x4-play-v0}  & $5$ & $3$  & $300$ & $(0.3, 0.01)$ & $3$  \\
    \bottomrule
    \end{tabular}
\end{table}

\begin{table}[h]
    \small
    \centering
    \caption{\footnotesize\textbf{Baseline hyperparameters on MMBench2.}
    Uncertainty penalty $\lambda$ for the model-based baselines, behavior-cloning
    coefficient $\alpha$ for FQL, and $(\alpha_{\mathrm{actor}}, \alpha_{\mathrm{critic}})$
    for ReBRAC, $\alpha$ for IQL.}
    \label{tab:baseline_hyperparameters_mmbench2}
    \begin{tabular}{llllll}
    \toprule
    \tt{Environment} &
    \tt{LOMPO} $(\lambda)$ & \tt{OffDV2} $(\lambda)$ &
    \tt{FQL} $(\alpha)$ & \tt{ReBRAC} $(\alpha_{\mathrm{actor}}, \alpha_{\mathrm{critic}})$ & \tt{IQL} $(\alpha)$ \\
    \midrule
    \texttt{cup-catch-var1}           & $1$   & $10$ & $0.3$ & $(0.01, 0)$ & $1$   \\
    \texttt{finger-turn-easy-var1}    & $30$  & $1$  & $0.3$ & $(0.01, 0.01)$ & $1$ \\
    \texttt{ms-push-banana}           & $30$  & $30$ & $1.0$ & $(1.0, 0)$ & $1$     \\
    \texttt{og-point-var1}            & N/A   & $10$ & $0.3$ & $(0.01, 0.01)$ & $1$ \\
    \texttt{og-point-var2}            & $1$   & $10$ & $0.3$ & $(0.01, 0.01)$ & $1$ \\
    \texttt{pygame-point-maze-var4}   & $3$   & $1$  & $0.1$ & $(0.01, 0)$ & $1$    \\
    \texttt{pygame-reacher-easy}      & $30$  & $1$  & $1.0$ & $(0.1, 0.01)$ & $1$  \\
    \texttt{pygame-dungeon-explorer1} & $10$  & $3$  & $0.3$ & $(0.1, 0)$  & $1$    \\
    \texttt{pygame-foraging}          & $30$  & $10$ & $0.1$ & $(1.0, 0)$  & $1$    \\
    \texttt{pygame-whirlpool}         & $3$   & $10$ & $0.1$ & $(0.1, 0)$  & $1$    \\
    \bottomrule
    \end{tabular}
\end{table}

\begin{table}[h]
    \small
    \centering
    \caption{\footnotesize\textbf{Baseline hyperparameter on Robomimic.} Uncertainty penalty $\lambda$ for the model-based baselines, action chunk size $n$ for BC, QC and QC-IFQL.}
    \label{tab:baseline_hyperparameters_robomimic_chunk}
    \begin{tabular}{llllll}
    \toprule
    \tt{Task} & \tt{LOMPO} $(\lambda)$ & \tt{OffDV2} $(\lambda)$ & \tt{BC} $(n)$ & \tt{QC} $(n)$ & \tt{QC-IFQL} $(n)$ \\
    \midrule
    \texttt{lift}      & $3$ & $10$ & $10$ & $5$ & $5$  \\
    \texttt{can}       & N/A & N/A  & $10$ & $5$ & $10$ \\
    \texttt{square}    & N/A & N/A  & $10$ & $5$ & $5$  \\
    \bottomrule
    \end{tabular}
\end{table}

\begin{table}[h]
    \small
    \centering
    \caption{\footnotesize\textbf{Per-task uncertainty coefficient $\lambda$ of the model-based baselines on LIBERO-10.} We sweep over $\lambda \in \{1, 3, 10, 30\}$ on each tasks for model-based baselines. N/A marks tasks on which every $\lambda$ scored zero.}
    \label{tab:baseline_hyperparameters_libero10}
    \begin{tabular}{p{10.0cm}ll}
    \toprule
    \tt{Task} & \tt{LOMPO} $(\lambda)$ & \tt{OffDV2} $(\lambda)$ \\
    \midrule
    \texttt{put the white mug on the left plate and put the yellow and white mug on the right plate} & N/A & N/A \\
    \texttt{put the white mug on the plate and put the chocolate pudding to the right of the plate} & N/A & N/A \\
    \texttt{put the yellow and white mug in the microwave and close it} & $3$ & N/A \\
    \texttt{turn on the stove and put the moka pot on it}               & N/A & N/A \\
    \texttt{put both the alphabet soup and the cream cheese box in the basket} & N/A & N/A \\
    \texttt{put both the alphabet soup and the tomato sauce in the basket} & $10$ & N/A \\
    \texttt{put both moka pots on the stove}                            & N/A & N/A \\
    \texttt{put both the cream cheese box and the butter in the basket}  & N/A & N/A \\
    \texttt{put the black bowl in the bottom drawer of the cabinet and close it} & N/A & N/A \\
    \texttt{pick up the book and place it in the back compartment of the caddy} & N/A & N/A \\
    \bottomrule
    \end{tabular}
\end{table}

\clearpage
\section{Complete Numerical Results}
\label{sec:number_results}
For completeness, we provide the full per-task results for OGBench tasks in \Cref{table:standard_full} (corresponding to \Cref{table:standard}). The results are averaged over $4$ seeds and we report the standard deviations for each tasks. We highlight the numbers with the best performance.

\begin{table}[h!]
\caption{
\footnotesize
\textbf{Complete results for OGBench visual benchmark tasks.}
}
\label{table:standard_full}

\vspace{-0.5em}
\centering
\scalebox{0.77}
{

\begin{tabular*}{1.3\textwidth}{p{5.2cm} R{0.5cm}@{\hspace{0.05cm}}L{0.33cm} R{0.5cm}@{\hspace{0.05cm}}L{0.33cm} R{0.5cm}@{\hspace{0.05cm}}L{0.33cm} R{0.5cm}@{\hspace{0.05cm}}L{0.33cm} R{0.5cm}@{\hspace{0.05cm}}L{0.33cm} R{0.5cm}@{\hspace{0.05cm}}L{0.33cm} R{0.5cm}@{\hspace{0.05cm}}L{0.33cm}
R{0.5cm}@{\hspace{0.05cm}}L{0.33cm}
R{0.5cm}@{\hspace{0.05cm}}L{0.33cm}}
\toprule
\multicolumn{1}{c}{} & \multicolumn{10}{c}{\tt{Model-Free}} & \multicolumn{6}{c}{\tt{Model-Based}} \\
\cmidrule(r){2-11}
\cmidrule(r){12-17}
\multicolumn{1}{l}{\tt{Environment}} & \multicolumn{2}{c}{\makebox[1.15cm]{\tt{IQL}}} & \multicolumn{2}{c}{\makebox[1.15cm]{\tt{ReBRAC}}} & \multicolumn{2}{c}{\makebox[1.15cm]{\tt{IFQL}}} & \multicolumn{2}{c}{\makebox[1.15cm]{\tt{FQL}}} & \multicolumn{2}{c}{\makebox[1.15cm]{\tt{QC}}} & \multicolumn{2}{c}{\makebox[1.15cm]{\tt{LOMPO}}} & \multicolumn{2}{c}{\makebox[1.15cm]{\tt{OffDV2}}} & \multicolumn{2}{c}{\makebox[1.15cm]{\color{myblue}\tt{\methodname}}} \\
\midrule
\tt{visual-cube-single-task1} & \tt{73} & {\tiny $\pm$\tt{7}} & \tt{84} & {\tiny $\pm$\tt{4}} & \tt{44} & {\tiny $\pm$\tt{6}} & \tt{67} & {\tiny $\pm$\tt{3}} & \tt{\color{myblue}{88}} & {\tiny $\pm$\tt{5}} & \tt{0} & {\tiny $\pm$\tt{0}} & \tt{0} & {\tiny $\pm$\tt{0}} & \tt{87} & {\tiny $\pm$\tt{2}} \\
\tt{visual-cube-single-task2} & \tt{57} & {\tiny $\pm$\tt{6}} & \tt{71} & {\tiny $\pm$\tt{6}} & \tt{29} & {\tiny $\pm$\tt{9}} & \tt{70} & {\tiny $\pm$\tt{4}} & \tt{\color{myblue}{80}} & {\tiny $\pm$\tt{5}} & \tt{0} & {\tiny $\pm$\tt{0}} & \tt{0} & {\tiny $\pm$\tt{0}} & \tt{72} & {\tiny $\pm$\tt{3}} \\
\tt{visual-cube-single-task3} & \tt{79} & {\tiny $\pm$\tt{4}} & \tt{85} & {\tiny $\pm$\tt{1}} & \tt{59} & {\tiny $\pm$\tt{10}} & \tt{68} & {\tiny $\pm$\tt{13}} & \tt{\color{myblue}{90}} & {\tiny $\pm$\tt{4}} & \tt{0} & {\tiny $\pm$\tt{0}} & \tt{0} & {\tiny $\pm$\tt{0}} & \tt{86} & {\tiny $\pm$\tt{3}} \\
\tt{visual-cube-single-task4} & \tt{62} & {\tiny $\pm$\tt{18}} & \tt{\color{myblue}{69}} & {\tiny $\pm$\tt{4}} & \tt{27} & {\tiny $\pm$\tt{7}} & \tt{51} & {\tiny $\pm$\tt{11}} & \tt{68} & {\tiny $\pm$\tt{5}} & \tt{0} & {\tiny $\pm$\tt{0}} & \tt{0} & {\tiny $\pm$\tt{0}} & \tt{64} & {\tiny $\pm$\tt{3}} \\
\tt{visual-cube-single-task5} & \tt{54} & {\tiny $\pm$\tt{4}} & \tt{\color{myblue}{62}} & {\tiny $\pm$\tt{4}} & \tt{22} & {\tiny $\pm$\tt{4}} & \tt{57} & {\tiny $\pm$\tt{8}} & \tt{\color{myblue}{62}} & {\tiny $\pm$\tt{6}} & \tt{0} & {\tiny $\pm$\tt{0}} & \tt{0} & {\tiny $\pm$\tt{0}} & \tt{\color{myblue}{62}} & {\tiny $\pm$\tt{3}} \\
\midrule
\tt{visual-cube-double-task1} & \tt{25} & {\tiny $\pm$\tt{7}} & \tt{5} & {\tiny $\pm$\tt{3}} & \tt{6} & {\tiny $\pm$\tt{2}} & \tt{15} & {\tiny $\pm$\tt{12}} & \tt{6} & {\tiny $\pm$\tt{2}} & \tt{0} & {\tiny $\pm$\tt{0}} & \tt{1} & {\tiny $\pm$\tt{1}} & \tt{\color{myblue}{67}} & {\tiny $\pm$\tt{4}} \\
\tt{visual-cube-double-task2} & \tt{4} & {\tiny $\pm$\tt{1}} & \tt{1} & {\tiny $\pm$\tt{0}} & \tt{0} & {\tiny $\pm$\tt{0}} & \tt{0} & {\tiny $\pm$\tt{0}} & \tt{2} & {\tiny $\pm$\tt{2}} & \tt{0} & {\tiny $\pm$\tt{0}} & \tt{0} & {\tiny $\pm$\tt{0}} & \tt{\color{myblue}{47}} & {\tiny $\pm$\tt{14}} \\
\tt{visual-cube-double-task3} & \tt{13} & {\tiny $\pm$\tt{4}} & \tt{4} & {\tiny $\pm$\tt{3}} & \tt{1} & {\tiny $\pm$\tt{1}} & \tt{2} & {\tiny $\pm$\tt{2}} & \tt{2} & {\tiny $\pm$\tt{1}} & \tt{0} & {\tiny $\pm$\tt{0}} & \tt{0} & {\tiny $\pm$\tt{0}} & \tt{\color{myblue}{52}} & {\tiny $\pm$\tt{6}} \\
\tt{visual-cube-double-task4} & \tt{1} & {\tiny $\pm$\tt{1}} & \tt{0} & {\tiny $\pm$\tt{0}} & \tt{0} & {\tiny $\pm$\tt{1}} & \tt{0} & {\tiny $\pm$\tt{0}} & \tt{0} & {\tiny $\pm$\tt{1}} & \tt{0} & {\tiny $\pm$\tt{0}} & \tt{0} & {\tiny $\pm$\tt{0}} & \tt{\color{myblue}{10}} & {\tiny $\pm$\tt{5}} \\
\tt{visual-cube-double-task5} & \tt{21} & {\tiny $\pm$\tt{3}} & \tt{10} & {\tiny $\pm$\tt{14}} & \tt{2} & {\tiny $\pm$\tt{1}} & \tt{4} & {\tiny $\pm$\tt{2}} & \tt{4} & {\tiny $\pm$\tt{1}} & \tt{0} & {\tiny $\pm$\tt{0}} & \tt{0} & {\tiny $\pm$\tt{0}} & \tt{\color{myblue}{90}} & {\tiny $\pm$\tt{4}} \\
\midrule
\tt{visual-scene-task1} & \tt{69} & {\tiny $\pm$\tt{40}} & \tt{97} & {\tiny $\pm$\tt{2}} & \tt{88} & {\tiny $\pm$\tt{4}} & \tt{99} & {\tiny $\pm$\tt{1}} & \tt{\color{myblue}{100}} & {\tiny $\pm$\tt{0}} & \tt{7} & {\tiny $\pm$\tt{5}} & \tt{9} & {\tiny $\pm$\tt{6}} & \tt{\color{myblue}{100}} & {\tiny $\pm$\tt{0}} \\
\tt{visual-scene-task2} & \tt{0} & {\tiny $\pm$\tt{0}} & \tt{65} & {\tiny $\pm$\tt{16}} & \tt{0} & {\tiny $\pm$\tt{1}} & \tt{88} & {\tiny $\pm$\tt{5}} & \tt{96} & {\tiny $\pm$\tt{3}} & \tt{5} & {\tiny $\pm$\tt{6}} & \tt{9} & {\tiny $\pm$\tt{5}} & \tt{\color{myblue}{100}} & {\tiny $\pm$\tt{0}} \\
\tt{visual-scene-task3} & \tt{17} & {\tiny $\pm$\tt{2}} & \tt{4} & {\tiny $\pm$\tt{3}} & \tt{15} & {\tiny $\pm$\tt{6}} & \tt{19} & {\tiny $\pm$\tt{5}} & \tt{89} & {\tiny $\pm$\tt{9}} & \tt{0} & {\tiny $\pm$\tt{0}} & \tt{0} & {\tiny $\pm$\tt{0}} & \tt{\color{myblue}{90}} & {\tiny $\pm$\tt{2}} \\
\tt{visual-scene-task4} & \tt{1} & {\tiny $\pm$\tt{0}} & \tt{0} & {\tiny $\pm$\tt{1}} & \tt{0} & {\tiny $\pm$\tt{0}} & \tt{1} & {\tiny $\pm$\tt{1}} & \tt{1} & {\tiny $\pm$\tt{1}} & \tt{0} & {\tiny $\pm$\tt{0}} & \tt{0} & {\tiny $\pm$\tt{0}} & \tt{\color{myblue}{42}} & {\tiny $\pm$\tt{29}} \\
\tt{visual-scene-task5} & \tt{0} & {\tiny $\pm$\tt{0}} & \tt{0} & {\tiny $\pm$\tt{0}} & \tt{0} & {\tiny $\pm$\tt{0}} & \tt{0} & {\tiny $\pm$\tt{0}} & \tt{0} & {\tiny $\pm$\tt{0}} & \tt{0} & {\tiny $\pm$\tt{0}} & \tt{0} & {\tiny $\pm$\tt{0}} & \tt{\color{myblue}{29}} & {\tiny $\pm$\tt{29}} \\
\midrule
\tt{visual-puzzle-3x3-task1} & \tt{7} & {\tiny $\pm$\tt{12}} & \tt{90} & {\tiny $\pm$\tt{4}} & \tt{\color{myblue}{100}} & {\tiny $\pm$\tt{1}} & \tt{97} & {\tiny $\pm$\tt{1}} & \tt{\color{myblue}{100}} & {\tiny $\pm$\tt{0}} & \tt{13} & {\tiny $\pm$\tt{9}} & \tt{9} & {\tiny $\pm$\tt{5}} & \tt{\color{myblue}{100}} & {\tiny $\pm$\tt{0}} \\
\tt{visual-puzzle-3x3-task2} & \tt{0} & {\tiny $\pm$\tt{0}} & \tt{0} & {\tiny $\pm$\tt{0}} & \tt{0} & {\tiny $\pm$\tt{0}} & \tt{0} & {\tiny $\pm$\tt{0}} & \tt{0} & {\tiny $\pm$\tt{0}} & \tt{0} & {\tiny $\pm$\tt{0}} & \tt{0} & {\tiny $\pm$\tt{0}} & \tt{\color{myblue}{100}} & {\tiny $\pm$\tt{0}} \\
\tt{visual-puzzle-3x3-task3} & \tt{0} & {\tiny $\pm$\tt{1}} & \tt{2} & {\tiny $\pm$\tt{1}} & \tt{1} & {\tiny $\pm$\tt{1}} & \tt{1} & {\tiny $\pm$\tt{0}} & \tt{0} & {\tiny $\pm$\tt{0}} & \tt{0} & {\tiny $\pm$\tt{0}} & \tt{0} & {\tiny $\pm$\tt{0}} & \tt{\color{myblue}{83}} & {\tiny $\pm$\tt{22}} \\
\tt{visual-puzzle-3x3-task4} & \tt{0} & {\tiny $\pm$\tt{0}} & \tt{1} & {\tiny $\pm$\tt{1}} & \tt{1} & {\tiny $\pm$\tt{0}} & \tt{6} & {\tiny $\pm$\tt{2}} & \tt{0} & {\tiny $\pm$\tt{0}} & \tt{0} & {\tiny $\pm$\tt{0}} & \tt{0} & {\tiny $\pm$\tt{0}} & \tt{\color{myblue}{98}} & {\tiny $\pm$\tt{2}} \\
\tt{visual-puzzle-3x3-task5} & \tt{1} & {\tiny $\pm$\tt{1}} & \tt{0} & {\tiny $\pm$\tt{1}} & \tt{1} & {\tiny $\pm$\tt{2}} & \tt{7} & {\tiny $\pm$\tt{3}} & \tt{0} & {\tiny $\pm$\tt{0}} & \tt{0} & {\tiny $\pm$\tt{0}} & \tt{0} & {\tiny $\pm$\tt{0}} & \tt{\color{myblue}{96}} & {\tiny $\pm$\tt{6}} \\
\midrule
\tt{visual-puzzle-4x4-task1} & \tt{0} & {\tiny $\pm$\tt{0}} & \tt{22} & {\tiny $\pm$\tt{5}} & \tt{11} & {\tiny $\pm$\tt{19}} & \tt{28} & {\tiny $\pm$\tt{5}} & \tt{57} & {\tiny $\pm$\tt{7}} & \tt{0} & {\tiny $\pm$\tt{0}} & \tt{2} & {\tiny $\pm$\tt{1}} & \tt{\color{myblue}{90}} & {\tiny $\pm$\tt{4}} \\
\tt{visual-puzzle-4x4-task2} & \tt{0} & {\tiny $\pm$\tt{0}} & \tt{9} & {\tiny $\pm$\tt{2}} & \tt{0} & {\tiny $\pm$\tt{1}} & \tt{1} & {\tiny $\pm$\tt{0}} & \tt{2} & {\tiny $\pm$\tt{1}} & \tt{0} & {\tiny $\pm$\tt{0}} & \tt{1} & {\tiny $\pm$\tt{0}} & \tt{\color{myblue}{21}} & {\tiny $\pm$\tt{8}} \\
\tt{visual-puzzle-4x4-task3} & \tt{0} & {\tiny $\pm$\tt{0}} & \tt{15} & {\tiny $\pm$\tt{2}} & \tt{2} & {\tiny $\pm$\tt{3}} & \tt{14} & {\tiny $\pm$\tt{3}} & \tt{31} & {\tiny $\pm$\tt{11}} & \tt{0} & {\tiny $\pm$\tt{0}} & \tt{1} & {\tiny $\pm$\tt{1}} & \tt{\color{myblue}{82}} & {\tiny $\pm$\tt{18}} \\
\tt{visual-puzzle-4x4-task4} & \tt{0} & {\tiny $\pm$\tt{0}} & \tt{10} & {\tiny $\pm$\tt{1}} & \tt{4} & {\tiny $\pm$\tt{4}} & \tt{7} & {\tiny $\pm$\tt{2}} & \tt{8} & {\tiny $\pm$\tt{8}} & \tt{0} & {\tiny $\pm$\tt{0}} & \tt{0} & {\tiny $\pm$\tt{1}} & \tt{\color{myblue}{20}} & {\tiny $\pm$\tt{7}} \\
\tt{visual-puzzle-4x4-task5} & \tt{0} & {\tiny $\pm$\tt{0}} & \tt{10} & {\tiny $\pm$\tt{1}} & \tt{0} & {\tiny $\pm$\tt{0}} & \tt{1} & {\tiny $\pm$\tt{0}} & \tt{1} & {\tiny $\pm$\tt{1}} & \tt{0} & {\tiny $\pm$\tt{0}} & \tt{1} & {\tiny $\pm$\tt{1}} & \tt{\color{myblue}{16}} & {\tiny $\pm$\tt{16}} \\
\bottomrule
\end{tabular*}
}
\vspace{-1.0em}
\end{table}

\begin{table}[h!]
\caption{
\footnotesize
\textbf{Complete results for MMBench2 tasks.}
}
\label{table:MMBench2_full}

\vspace{-0.5em}
\centering
\scalebox{0.77}
{

\begin{tabular*}{1.3\textwidth}{p{5.2cm} R{0.5cm}@{\hspace{0.05cm}}L{0.33cm} R{0.5cm}@{\hspace{0.05cm}}L{0.33cm} R{0.5cm}@{\hspace{0.05cm}}L{0.33cm} R{0.5cm}@{\hspace{0.05cm}}L{0.33cm} R{0.5cm}@{\hspace{0.05cm}}L{0.33cm} R{0.5cm}@{\hspace{0.05cm}}L{0.33cm} R{0.5cm}@{\hspace{0.05cm}}L{0.33cm}
R{0.5cm}@{\hspace{0.05cm}}L{0.33cm}}
\toprule
\multicolumn{1}{c}{} & \multicolumn{10}{c}{\tt{Model-Free}} & \multicolumn{6}{c}{\tt{Model-Based}} \\
\cmidrule(r){2-11}
\cmidrule(r){12-17}
\multicolumn{1}{l}{\tt{Environment}} & \multicolumn{2}{c}{\makebox[1.15cm]{\tt{IQL}}} & \multicolumn{2}{c}{\makebox[1.15cm]{\tt{ReBRAC}}} & \multicolumn{2}{c}{\makebox[1.15cm]{\tt{IFQL}}} & \multicolumn{2}{c}{\makebox[1.15cm]{\tt{FQL}}} & \multicolumn{2}{c}{\makebox[1.15cm]{\tt{QC}}} & \multicolumn{2}{c}{\makebox[1.15cm]{\tt{LOMPO}}} & \multicolumn{2}{c}{\makebox[1.15cm]{\tt{OffDV2}}} & \multicolumn{2}{c}{\makebox[1.15cm]{\color{myblue}\tt{\methodname}}} \\
\midrule
\tt{cup-catch-var1} & \tt{13} & {\tiny $\pm$\tt{2}} & \tt{47} & {\tiny $\pm$\tt{6}} & \tt{11} & {\tiny $\pm$\tt{1}} & \tt{\color{myblue}{71}} & {\tiny $\pm$\tt{1}} & \tt{22} & {\tiny $\pm$\tt{4}} & \tt{4} & {\tiny $\pm$\tt{3}} & \tt{45} & {\tiny $\pm$\tt{26}} & \tt{33} & {\tiny $\pm$\tt{4}} \\
\tt{finger-turn-easy-var1} & \tt{24} & {\tiny $\pm$\tt{5}} & \tt{32} & {\tiny $\pm$\tt{4}} & \tt{23} & {\tiny $\pm$\tt{4}} & \tt{35} & {\tiny $\pm$\tt{3}} & \tt{23} & {\tiny $\pm$\tt{2}} & \tt{24} & {\tiny $\pm$\tt{9}} & \tt{\color{myblue}{43}} & {\tiny $\pm$\tt{1}} & \tt{29} & {\tiny $\pm$\tt{5}} \\
\tt{ms-push-banana} & \tt{5} & {\tiny $\pm$\tt{2}} & \tt{8} & {\tiny $\pm$\tt{3}} & \tt{11} & {\tiny $\pm$\tt{4}} & \tt{3} & {\tiny $\pm$\tt{2}} & \tt{3} & {\tiny $\pm$\tt{2}} & \tt{\color{myblue}{23}} & {\tiny $\pm$\tt{3}} & \tt{0} & {\tiny $\pm$\tt{0}} & \tt{4} & {\tiny $\pm$\tt{2}} \\
\tt{og-point-var1} & \tt{8} & {\tiny $\pm$\tt{5}} & \tt{17} & {\tiny $\pm$\tt{2}} & \tt{11} & {\tiny $\pm$\tt{3}} & \tt{32} & {\tiny $\pm$\tt{6}} & \tt{7} & {\tiny $\pm$\tt{2}} & \tt{0} & {\tiny $\pm$\tt{0}} & \tt{1} & {\tiny $\pm$\tt{0}} & \tt{\color{myblue}{42}} & {\tiny $\pm$\tt{5}} \\
\tt{og-point-var2} & \tt{7} & {\tiny $\pm$\tt{4}} & \tt{14} & {\tiny $\pm$\tt{3}} & \tt{9} & {\tiny $\pm$\tt{3}} & \tt{\color{myblue}{28}} & {\tiny $\pm$\tt{5}} & \tt{6} & {\tiny $\pm$\tt{2}} & \tt{0} & {\tiny $\pm$\tt{0}} & \tt{1} & {\tiny $\pm$\tt{0}} & \tt{24} & {\tiny $\pm$\tt{4}} \\
\tt{pygame-point-maze-var4} & \tt{10} & {\tiny $\pm$\tt{6}} & \tt{25} & {\tiny $\pm$\tt{14}} & \tt{26} & {\tiny $\pm$\tt{7}} & \tt{45} & {\tiny $\pm$\tt{4}} & \tt{16} & {\tiny $\pm$\tt{1}} & \tt{2} & {\tiny $\pm$\tt{3}} & \tt{5} & {\tiny $\pm$\tt{1}} & \tt{\color{myblue}{50}} & {\tiny $\pm$\tt{8}} \\
\tt{pygame-reacher-easy} & \tt{11} & {\tiny $\pm$\tt{2}} & \tt{11} & {\tiny $\pm$\tt{1}} & \tt{10} & {\tiny $\pm$\tt{1}} & \tt{12} & {\tiny $\pm$\tt{2}} & \tt{9} & {\tiny $\pm$\tt{1}} & \tt{5} & {\tiny $\pm$\tt{1}} & \tt{\color{myblue}{28}} & {\tiny $\pm$\tt{4}} & \tt{22} & {\tiny $\pm$\tt{3}} \\
\tt{pygame-dungeon-explorer1} & \tt{43} & {\tiny $\pm$\tt{2}} & \tt{57} & {\tiny $\pm$\tt{4}} & \tt{46} & {\tiny $\pm$\tt{3}} & \tt{65} & {\tiny $\pm$\tt{5}} & \tt{64} & {\tiny $\pm$\tt{1}} & \tt{27} & {\tiny $\pm$\tt{8}} & \tt{28} & {\tiny $\pm$\tt{7}} & \tt{\color{myblue}{78}} & {\tiny $\pm$\tt{3}} \\
\tt{pygame-foraging} & \tt{65} & {\tiny $\pm$\tt{0}} & \tt{65} & {\tiny $\pm$\tt{2}} & \tt{65} & {\tiny $\pm$\tt{1}} & \tt{76} & {\tiny $\pm$\tt{1}} & \tt{68} & {\tiny $\pm$\tt{1}} & \tt{44} & {\tiny $\pm$\tt{8}} & \tt{52} & {\tiny $\pm$\tt{3}} & \tt{\color{myblue}{82}} & {\tiny $\pm$\tt{1}} \\
\tt{pygame-whirlpool} & \tt{74} & {\tiny $\pm$\tt{7}} & \tt{89} & {\tiny $\pm$\tt{8}} & \tt{86} & {\tiny $\pm$\tt{4}} & \tt{91} & {\tiny $\pm$\tt{1}} & \tt{92} & {\tiny $\pm$\tt{3}} & \tt{79} & {\tiny $\pm$\tt{18}} & \tt{\color{myblue}{99}} & {\tiny $\pm$\tt{1}} & \tt{\color{myblue}{99}} & {\tiny $\pm$\tt{1}} \\
\bottomrule
\end{tabular*}
}
\vspace{-1.0em}
\end{table}

\begin{table}[t!]
\caption{
\footnotesize
\textbf{Complete results for LIBERO tasks.}
}
\label{table:robotics_full}

\vspace{-0.5em}
\centering
\scalebox{0.76}
{

\begin{tabular*}{1.31\textwidth}{p{8.5cm} R{0.5cm}@{\hspace{0.05cm}}L{0.33cm} R{0.5cm}@{\hspace{0.05cm}}L{0.33cm} R{0.5cm}@{\hspace{0.05cm}}L{0.33cm} 
R{0.5cm}@{\hspace{0.05cm}}L{0.33cm} 
R{0.5cm}@{\hspace{0.05cm}}L{0.33cm} R{0.5cm}@{\hspace{0.05cm}}L{0.33cm} }
\toprule
\multicolumn{1}{c}{} & \multicolumn{6}{c}{\tt{Model-Free}} & \multicolumn{6}{c}{\tt{Model-Based}} \\
\cmidrule(r){2-7}
\cmidrule(r){8-13}
\multicolumn{1}{l}{\tt{Environment}} & \multicolumn{2}{c}{\makebox[1.15cm]{\tt{BC}}} & \multicolumn{2}{c}{\makebox[1.15cm]{\tt{QC}}} & \multicolumn{2}{c}{\makebox[1.15cm]{\tt{QC-IFQL}}} & \multicolumn{2}{c}
{\makebox[1.15cm]{\tt{LOMPO}}} & \multicolumn{2}{c}
{\makebox[1.15cm]{\tt{OffDV2}}} & \multicolumn{2}{c}{\makebox[1.15cm]{\color{myblue}\tt{\methodname}}} \\
\midrule
\multicolumn{9}{l}{\texttt{LIBERO (10 tasks)}}\\
\midrule
\tt{turn on the stove and put the moka pot on it} & \tt{96} & {\tiny $\pm$\tt{3}} & \tt{89} & {\tiny $\pm$\tt{10}} & \tt{85} & {\tiny $\pm$\tt{9}} & \tt{0} & {\tiny $\pm$\tt{0}} & \tt{0} & {\tiny $\pm$\tt{0}} & \tt{\color{myblue}{97}} & {\tiny $\pm$\tt{2}} \\
\tt{put the black bowl in the bottom drawer of the cabinet and close it} & \tt{\color{myblue}{98}} & {\tiny $\pm$\tt{1}} & \tt{86} & {\tiny $\pm$\tt{9}} & \tt{94} & {\tiny $\pm$\tt{3}} & \tt{0} & {\tiny $\pm$\tt{0}} & \tt{0} & {\tiny $\pm$\tt{0}} & \tt{95} & {\tiny $\pm$\tt{3}} \\
\tt{put the yellow and white mug in the microwave and close it} & \tt{\color{myblue}{84}} & {\tiny $\pm$\tt{3}} & \tt{77} & {\tiny $\pm$\tt{7}} & \tt{83} & {\tiny $\pm$\tt{8}} & \tt{9} & {\tiny $\pm$\tt{12}} & \tt{0} & {\tiny $\pm$\tt{0}} & \tt{81} & {\tiny $\pm$\tt{8}} \\
\tt{put both moka pots on the stove} & \tt{\color{myblue}{76}} & {\tiny $\pm$\tt{4}} & \tt{69} & {\tiny $\pm$\tt{2}} & \tt{64} & {\tiny $\pm$\tt{10}} & \tt{0} & {\tiny $\pm$\tt{0}} & \tt{0} & {\tiny $\pm$\tt{0}} & \tt{75} & {\tiny $\pm$\tt{16}} \\
\tt{put both the alphabet soup and the cream cheese box in the basket} & \tt{69} & {\tiny $\pm$\tt{4}} & \tt{67} & {\tiny $\pm$\tt{18}} & \tt{68} & {\tiny $\pm$\tt{8}} & \tt{0} & {\tiny $\pm$\tt{0}} & \tt{0} & {\tiny $\pm$\tt{0}} & \tt{\color{myblue}{75}} & {\tiny $\pm$\tt{6}} \\
\tt{put both the alphabet soup and the tomato sauce in the basket} & \tt{77} & {\tiny $\pm$\tt{5}} & \tt{80} & {\tiny $\pm$\tt{8}} & \tt{75} & {\tiny $\pm$\tt{8}} & \tt{2} & {\tiny $\pm$\tt{4}} & \tt{0} & {\tiny $\pm$\tt{0}} & \tt{\color{myblue}{82}} & {\tiny $\pm$\tt{7}} \\
\tt{put both the cream cheese box and the butter in the basket} & \tt{84} & {\tiny $\pm$\tt{10}} & \tt{\color{myblue}{90}} & {\tiny $\pm$\tt{6}} & \tt{85} & {\tiny $\pm$\tt{3}} & \tt{0} & {\tiny $\pm$\tt{0}} & \tt{0} & {\tiny $\pm$\tt{0}} & \tt{86} & {\tiny $\pm$\tt{5}} \\
\tt{put the white mug on the left plate and put the yellow and white mug on the right plate} & \tt{\color{myblue}{83}} & {\tiny $\pm$\tt{6}} & \tt{71} & {\tiny $\pm$\tt{11}} & \tt{\color{myblue}{83}} & {\tiny $\pm$\tt{3}} & \tt{0} & {\tiny $\pm$\tt{0}} & \tt{0} & {\tiny $\pm$\tt{0}} & \tt{74} & {\tiny $\pm$\tt{9}} \\
\tt{put the white mug on the plate and put the chocolate pudding to the right of the plate} & \tt{49} & {\tiny $\pm$\tt{7}} & \tt{59} & {\tiny $\pm$\tt{5}} & \tt{\color{myblue}{68}} & {\tiny $\pm$\tt{14}} & \tt{0} & {\tiny $\pm$\tt{0}} & \tt{0} & {\tiny $\pm$\tt{0}} & \tt{48} & {\tiny $\pm$\tt{9}} \\
\tt{pick up the book and place it in the back compartment of the caddy} & \tt{\color{myblue}{84}} & {\tiny $\pm$\tt{5}} & \tt{80} & {\tiny $\pm$\tt{5}} & \tt{78} & {\tiny $\pm$\tt{7}} & \tt{0} & {\tiny $\pm$\tt{0}} & \tt{0} & {\tiny $\pm$\tt{0}} & \tt{82} & {\tiny $\pm$\tt{5}} \\

\bottomrule
\end{tabular*}
}
\vspace{-1.0em}
\end{table}

\clearpage
\section{Training Curves}
\begin{figure}[h!]
    \centering
    \includegraphics[width=1.0\textwidth]{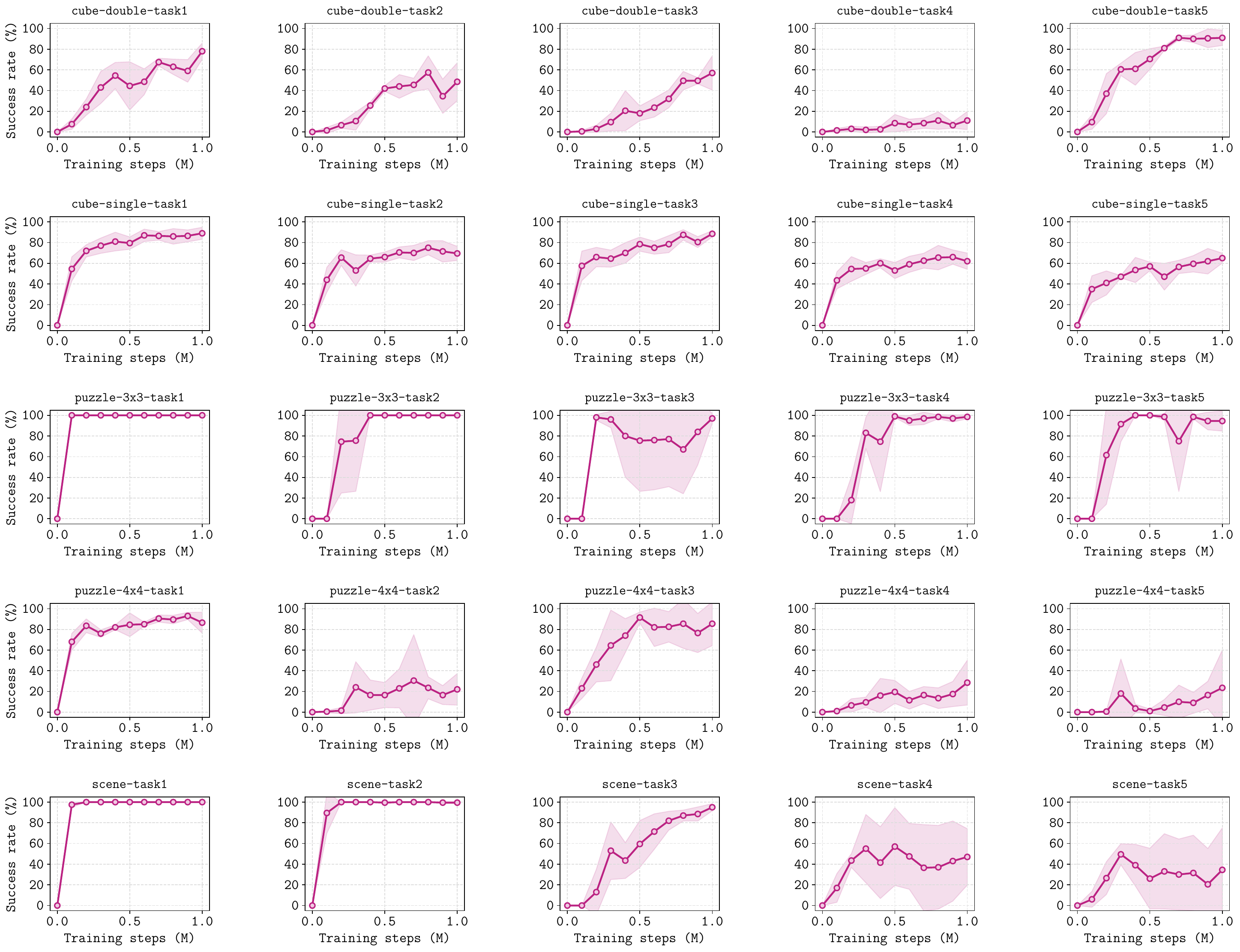}
    \caption{
    \footnotesize
    \textbf{Training curve of \methodname in OGBench environments.} We report the success rate for 15 evaluation episodes across 4 seeds (total 60 episodes). Shaded region represents the [mean - std, mean + std].
    }
    \label{fig:training_curves_ogbench}
\end{figure}

\begin{figure}[h!]
    \centering
    \includegraphics[width=1.0\textwidth]{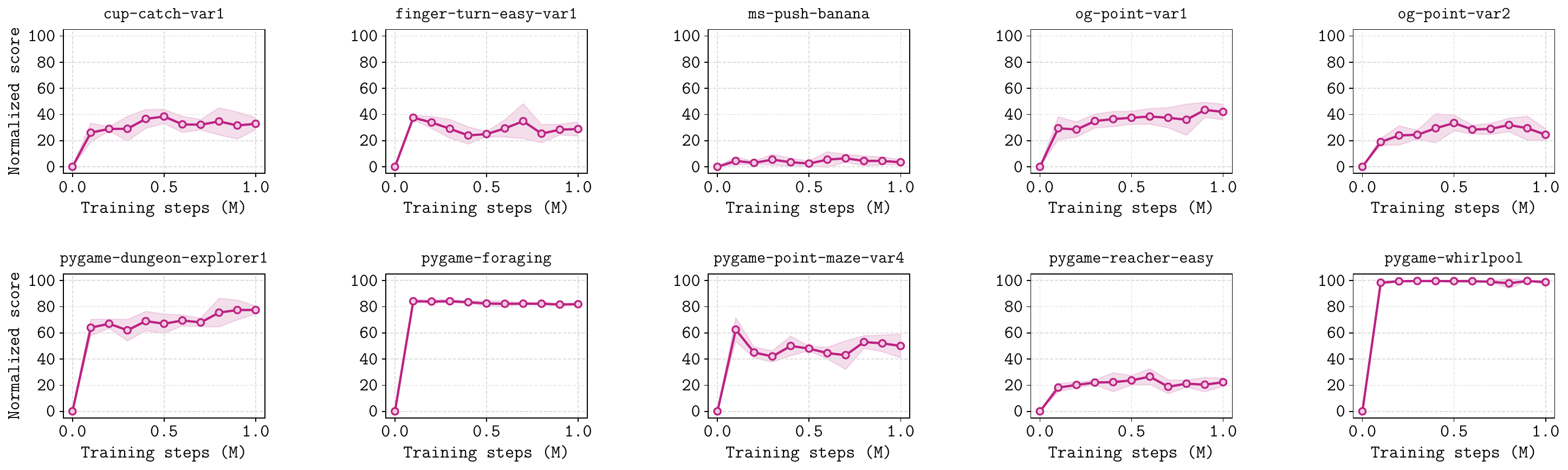}
    \caption{
    \footnotesize
    \textbf{Training curve of \methodname in MMBench2 environments.} We report the normalized return for 15 evaluation episodes across 4 seeds (total 60 episodes). Shaded region represents the [mean - std, mean + std].
    }
    \label{fig:training_curves_MMBench2}
\end{figure}

\begin{figure}[h!]
    \centering
    \includegraphics[width=0.87\textwidth]{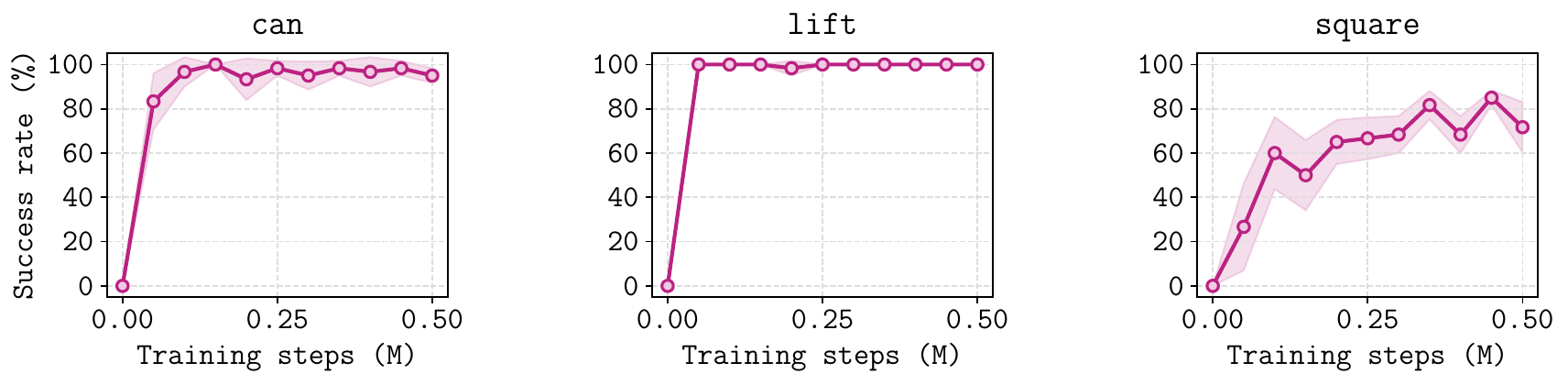}
    \caption{
    \footnotesize
    \textbf{Training curve of \methodname in Robomimic environments.} We report the success rate for 15 evaluation episodes across 4 seeds (total 60 episodes). Shaded region represents the [mean - std, mean + std].
    }
    \label{fig:training_curves_robomimic}
\end{figure}

\begin{figure}[h!]
    \centering
    \includegraphics[width=1.0\textwidth]{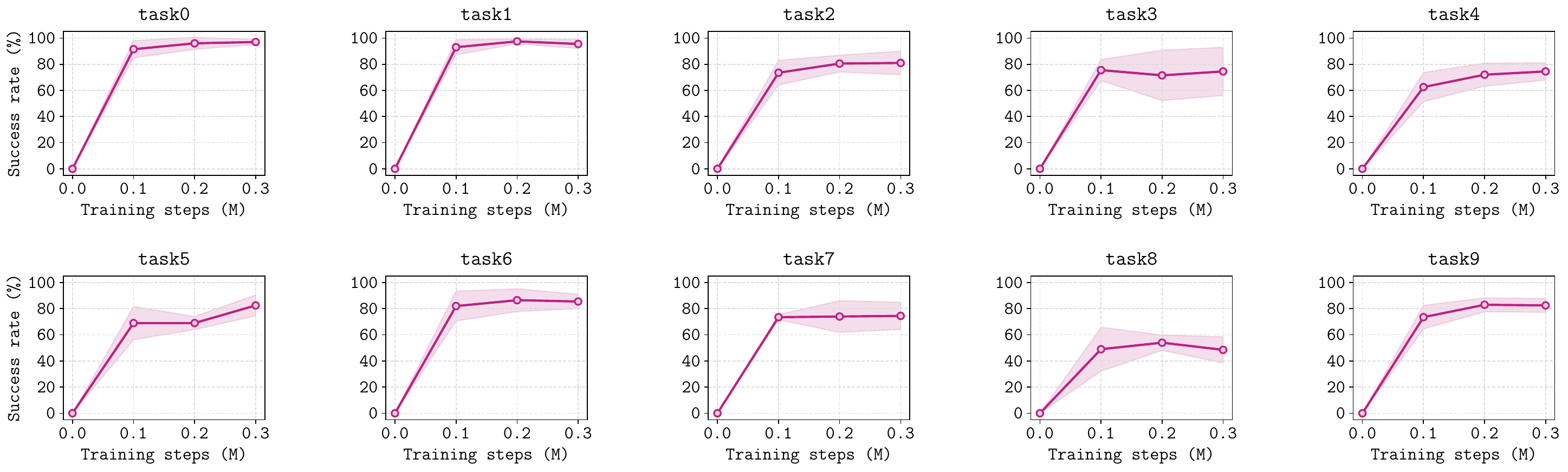}
    \caption{
    \footnotesize
    \textbf{Training curve of \methodname in LIBERO-10 environments.} We report the success rate for 15 evaluation episodes across 4 seeds (total 60 episodes). Shaded region represents the [mean - std, mean + std]. For the task instruction of each tasks for this graph, please refer to \Cref{table:robotics}.
    }
    \label{fig:training_curves_libero}
\end{figure}

We provide the training curve of \methodname for the OGBench, MMBench, robomimic, LIBERO-10 result of \Cref{table:standard,table:robotics}, in \Cref{fig:training_curves_ogbench,fig:training_curves_MMBench2,fig:training_curves_robomimic,fig:training_curves_libero}. 
We plot the mean and the standard deviation (across 4 seeds) by  covering [mean - std, mean + std] area with a lighter color.

\end{document}